\documentclass[runningheads]{llncs}

\usepackage{eccv}

\usepackage{eccvabbrv}
\usepackage{multibib}
\usepackage{multicol}
\newcites{supp}{Supplementary References}
\usepackage{graphicx}
\usepackage{booktabs}
\usepackage{siunitx}
\usepackage{colortbl}
\usepackage{xcolor}
\usepackage{adjustbox}
\usepackage{multirow}
\usepackage{mathtools}
\usepackage{algorithm}
\usepackage{algpseudocode}
\usepackage{enumitem}
\usepackage{makecell}
\definecolor{bestblue}{RGB}{135, 206, 250}      
\definecolor{secondblue}{RGB}{224, 240, 255}    

\newcommand{\best}[1]{\cellcolor{bestblue}{\textbf{#1}}}

\newcommand{\second}[1]{\cellcolor{secondblue}{#1}}
\usepackage[accsupp]{axessibility}  

\usepackage{hyperref}
\hypersetup{
pdftitle={UniTriSplat: A Unified 3D Gaussian Splatting Framework with Uniform Spherical Rasterization for Universal Cameras},
pdfauthor={Yipeng Zhu, Huajian Huang, Tristan Braud, Sai-Kit Yeung},
pdfkeywords={3D Gaussian Splatting, omnidirectional reconstruction, novel view synthesis}
}
\usepackage{orcidlink}

\begin{document}

\title{UniTriSplat: A Unified 3D Gaussian Splatting Framework with Uniform Spherical Rasterization for Universal Cameras}

\author{Yipeng Zhu\inst{1}\orcidlink{0009-0002-9984-2269} \and
Huajian Huang\inst{2}$^\dagger$\orcidlink{0000-0002-0963-1146} \and
Tristan Braud\inst{1}\orcidlink{0000-0002-9571-0544} \and
Sai-Kit Yeung\inst{1}\orcidlink{0000-0001-7974-0607} }

\authorrunning{Y. Zhu et al.}
\titlerunning{UniTriSplat}

\institute{The Hong Kong University of Science and Technology, Hong Kong, China \and
Beijing Institute of Technology, Beijing, China\\
  $^\dagger$Corresponding author}

\maketitle

\begin{abstract}
Existing 3D Gaussian Splatting (3DGS) frameworks rely on camera-specific rasterization, suffering from inconsistent solid-angle sampling and degraded performance across heterogeneous camera models (e.g., perspective, fisheye, omnidirectional).
To address this limitation, we propose UniTriSplat, a unified 3DGS framework for universal cameras that reformulates Gaussian splatting on the unit sphere via HEALPix discretization.
Leveraging the equal-area property of HEALPix, we construct a spherical sampling grid aligned with the angular resolution of input images. We derive the forward rendering and gradient propagation of Gaussians directly in the spherical radian domain, yielding uniform optimization behavior from narrow-FoV images to full 360-degree panoramas.
To enhance perceptual reconstruction quality, we additionally introduce a HEALPix-aware SSIM loss that respects spherical neighborhood structure. 
Extensive experiments across diverse camera models demonstrate that UniTriSplat consistently improves cross-camera generalization while preserving geometric fidelity and rendering quality. Project page: https://yipengzhu0809.github.io/UniTriSplat/

\keywords{3D Gaussian Splatting \and Omnidirectional Reconstruction \and Novel View Synthesis}
\end{abstract}
\section{Introduction}

3D Gaussian Splatting~\cite{3dgs} (3DGS) achieves high-fidelity 3D reconstruction by employing explicit primitives and a differentiable tile-based rasterization pipeline. 
However, increasing demand for wide-field-of-view (FoV) perception in urban-scale digital twins~\cite{hugs, street} and immersive virtual reality~\cite{vrgs,vrsplat} requires a shift from perspective rendering to omnidirectional synthesis. 
This transition necessitates that 3DGS natively supports heterogeneous imaging models across arbitrary-FoV configurations, including fisheye lenses and \(360^\circ\) imagery.

\begin{figure}[!t]
    \centering
    \includegraphics[width=0.8\textwidth]{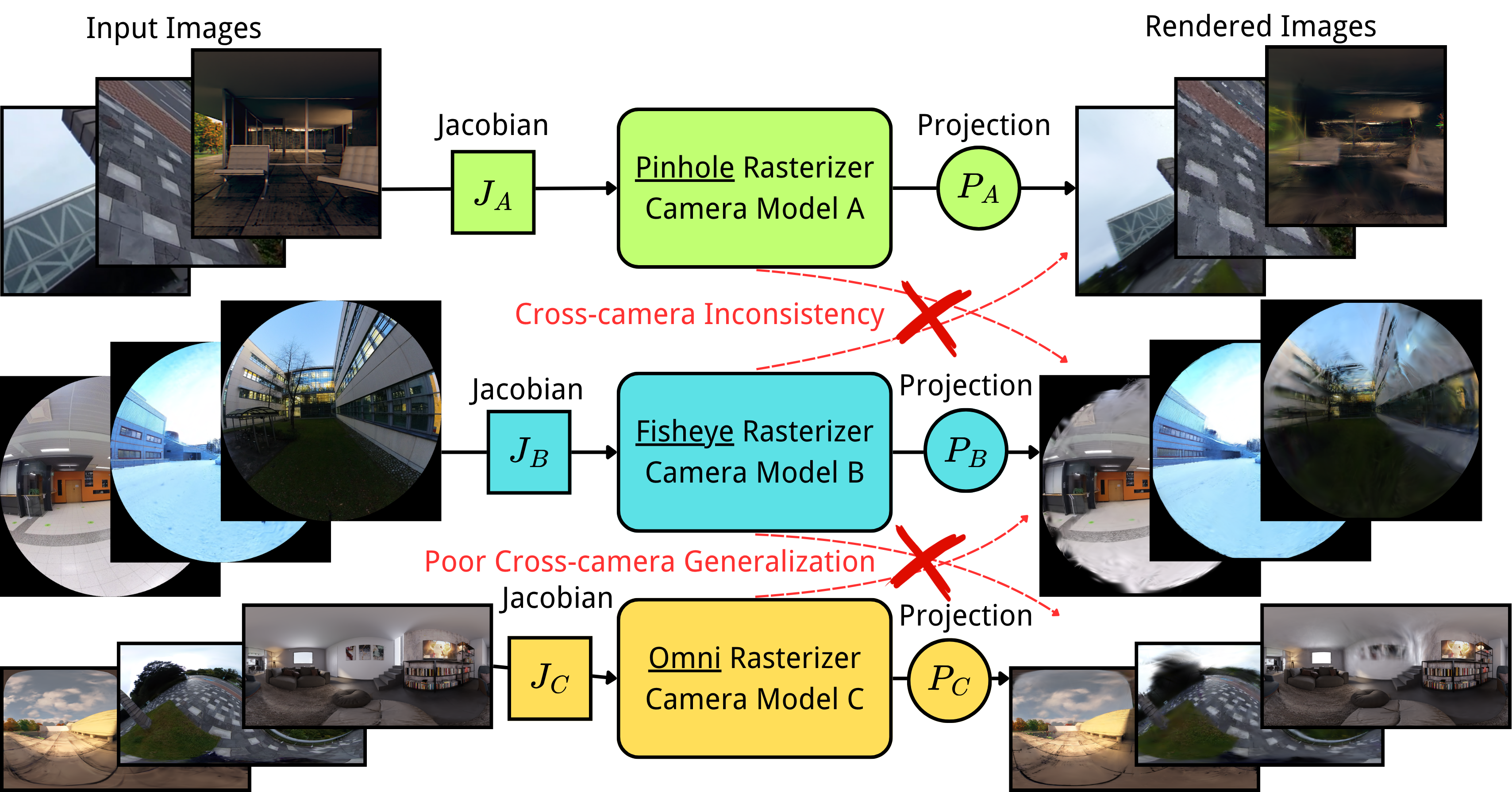} 
    \caption{\textbf{Limitations of camera-specific 3DGS pipelines.} Existing methods require separate rasterizers and projection models for different camera types, resulting in fragmented implementations and poor cross-camera generalization.}
    \label{fig:crossmodel} 
\end{figure}

Despite its efficiency, adapting 3DGS to diverse camera geometries remains challenging. 
Existing frameworks typically rely on camera-specific projection functions~\cite{fisheyegs} and their corresponding Jacobian matrices~\cite{ewa}, which leads to fragmented implementations and limits their versatility.
Crucially, the tight coupling between the rasterizer and the pinhole model prevents the learned Gaussian scene from maintaining geometric consistency when rendered across different camera types, thereby compromising cross-camera generalization (\cref{fig:crossmodel}).
An alternative is to map diverse camera models onto a unified intermediate space, such as equirectangular panoramas~\cite{omnigs, op43dgs, environment}.  
Nevertheless, this approach is limited by the requirement for $360^\circ$ coverage and struggles with partial-FoV inputs. 
Furthermore, the non-uniform pixel density of equirectangular projection induces imbalanced sampling between polar and equatorial zones~\cite{erpgs}, resulting in biased gradient propagation and suboptimal convergence (\cref{fig:rasterization}).

Fundamentally, all central camera models can be unified through a shared spherical geometry.
In this work, we exploit this principle to develop \textbf{UniTriSplat}, a unified 3DGS framework featuring a camera-agnostic spherical rasterizer. 
By projecting 3D Gaussians directly onto a standardized spherical grid, UniTriSplat decouples the rendering process from camera-specific projection, enabling consistent synthesis across arbitrary FoVs and lens models. 
Specifically, we adopt HEALPix~\cite{healpix} (Hierarchical Equal Area iso-Latitude Pixelization) as the underlying spherical discretization scheme. 
HEALPix partitions the sphere into a hierarchical grid of equal-area pixels, inherently eliminating the sampling bias present in equirectangular representations~\cite{healswin}.
To accommodate varying camera configurations, we adjust the HEALPix resolution by matching the grid's solid angle to the local angular resolution of the input images.
Building upon this representation, we reformulate the forward and backward rasterization of 3DGS within the arc-length coordinate system on the unit sphere, and employ HEALPix for uniform spherical tessellation. 
Subsequently, we redesign the tile query, depth sorting, and density control modules to operate natively on the HEALPix grid. 
In particular, the tile query supports both sequential scanning and quadtree traversal via different indexing schemes, integrating depth sorting by radial distance and density control driven by angular-gradient thresholds.
During rendering, the 3D Gaussian scene is rasterized onto a spherical domain corresponding to the target FoV, followed by resampling onto the image plane of the target camera. 
To further extend spherical perceptual quality~\cite{scomnigs} to HEALPix, we introduce a HEALPix-aware SSIM (Structural Similarity Index) loss that computes structural similarity over spherical neighborhoods, improving reconstruction fidelity in distorted regions.
The pipeline of UniTriSplat is illustrated in~\cref{fig:pipeline}. Our contributions can be summarized as follows:

\begin{figure}[!t]
    \centering
    \includegraphics[width=1.0\textwidth]{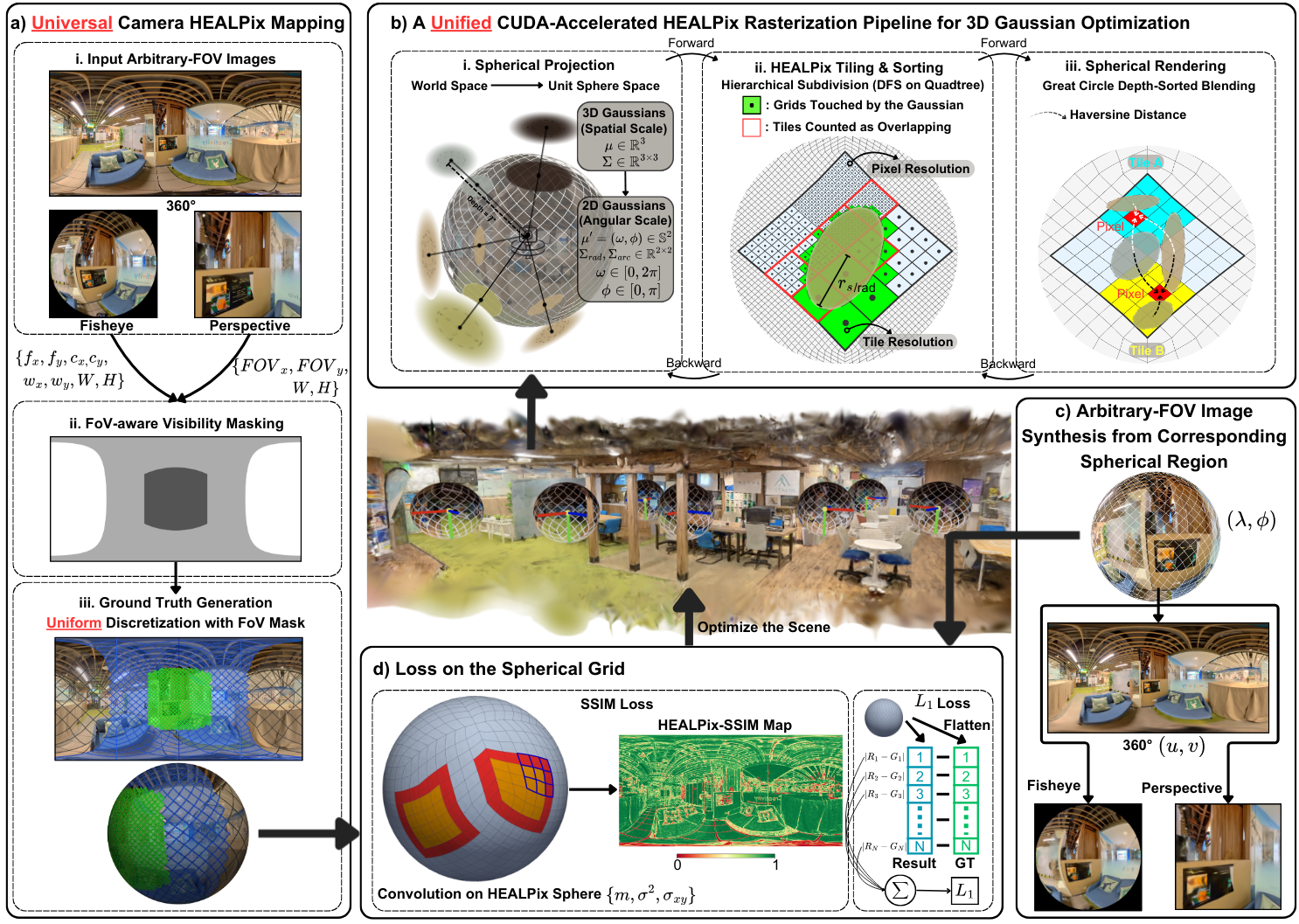} 
    \caption{\textbf{Overview of UniTriSplat.} (a) Unified mapping from heterogeneous camera models to the HEALPix domain with FoV-specific masking. (b) Spherical rasterization within the radian space: projection, tile query, and depth-sorted blending.  (c) Sampling the rendered HEALPix grids into  2D images for evaluation. (d) Optimization guided by the HEALPix-aware SSIM and $L_1$ losses.}
    \label{fig:pipeline} 
\end{figure}

\begin{itemize}
    \item Proposing UniTriSplat, the first 3DGS framework supporting arbitrary camera models and FoVs via a unified HEALPix-based rasterization pipeline.
    \item Deriving the gradients for splatting 3D Gaussians onto HEALPix spherical grids and developing a custom rasterizer for effective training and rendering. 
    \item Designing a HEALPix-SSIM loss for spherical geometry in structural similarity computation, facilitating spatially uniform Gaussian optimization.
    \item Comprehensive experiments demonstrating state-of-the-art spherical reconstruction metrics, competitive planar metrics, improved cross-camera generalization, and stable reconstruction quality for arbitrary-FoV inputs.
\end{itemize}

\section{Related Work}
\label{sec:relatedwork}

\subsection{3D Gaussian Splatting Beyond Perspective Projection}
Wide-FoV cameras are essential for VR, robotics, and autonomous systems, yet vanilla 3DGS relies on perspective projection, introducing severe distortions for such inputs (\cref{fig:rasterization}). 
Recent works~\cite{fisheyegs,omnigs} derive camera-specific Jacobians for equidistant and equirectangular projections.
Others employ tangent-plane projections~\cite{360gs,op43dgs,odgs}, splatting Gaussians onto local tangent planes then warping to the sphere, though this introduces linearization artifacts at patch boundaries.
Several works~\cite{op43dgs,spags,3dgut} redesign the EWA affine approximation~\cite{ewa} for wide-FoV imaging; notably, Wu et al.~\cite{3dgut} replace it with the Unscented Transform for more accurate nonlinear projection under arbitrary distortions.
Ito et al.~\cite{erpgs} apply distortion-aware reweighting to mitigate polar oversampling, while Shin et al.~\cite{seam360gs} jointly optimize with dual-fisheye calibration for seamless $360^\circ$ renderings.
Deng et al.~\cite{deng2025self} propose a cross-FoV method that extends perspective projection to a cubemap representation, enabling the processing of large-FoV images.
However, all these methods require camera-specific modifications and remain confined to equirectangular or planar rasterization, leaving nonuniform sampling unaddressed. 
Our work performs rasterization directly on the sphere, providing a camera-agnostic framework with inherently uniform coverage.

\subsection{Spherical Image Representation}

\begin{figure}[!t]
    \centering
    \includegraphics[width=1.0\textwidth]{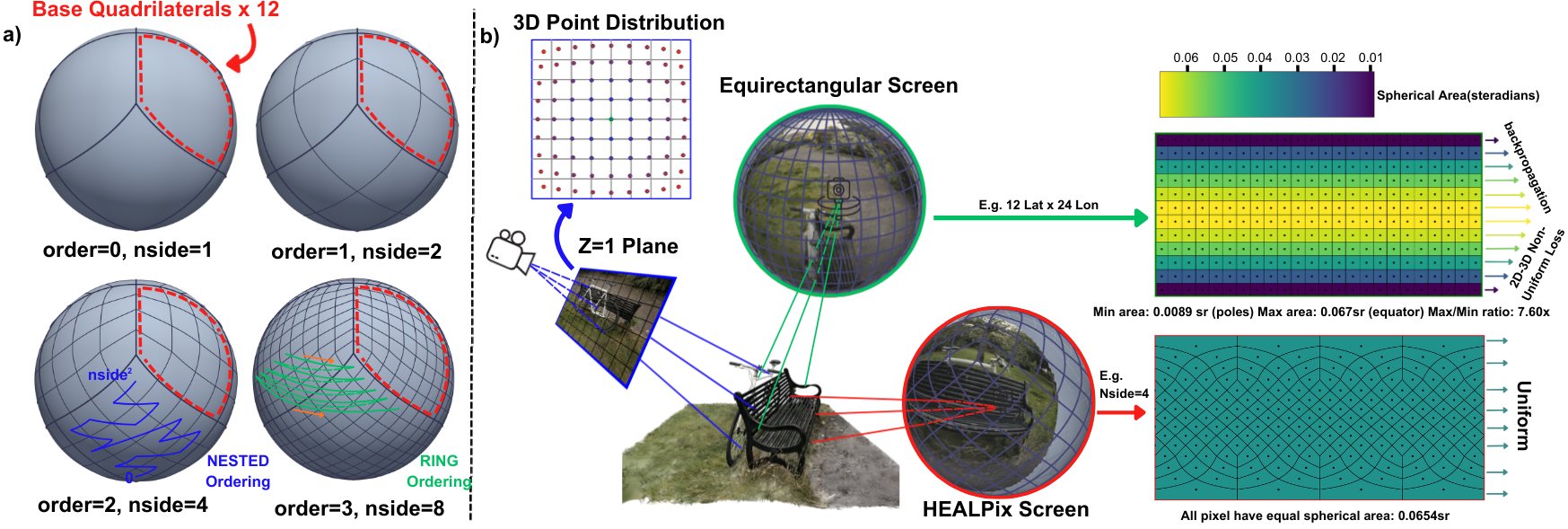} 
    \caption{\textbf{HEALPix Tessellation and Uniform Rasterization.} (a) Hierarchical structure and indexing. (b) Equirectangular projection shows $7.6\times$ pixel area variation, while HEALPix ensures uniform spherical sampling.}
    \label{fig:rasterization} 
\end{figure}

Projecting visual signals onto a sphere aligns naturally with omnidirectional perception. 
Since omnidirectional imaging was widely adopted~\cite{quicktime}, equirectangular projection~\cite{mapping} has become popular for its simplicity, supporting various vision tasks~\cite{360dvo,360vot,360loc,spherenet,distortioncnn}, but suffers from severe polar oversampling. 
To address this, uniform tessellation schemes have been explored~\cite{sphericaltransformer}, including cubemap~\cite{environment}, icosahedral grids~\cite{icosahedral-hexagonal}, and quasi-uniform Yin–Yang grid~\cite{yinyang}. 
Choi et al.~\cite{egonerf} use Yin-Yang grid for balanced ray sampling in scene reconstruction, while Lee et al.~\cite{omnisplat} leverage it for quasi-uniform patch decomposition enabling pretrained networks on $360^\circ$ images.
We adopt HEALPix as our spherical representation (\cref{fig:rasterization}). 
Unlike quasi-uniform schemes, HEALPix guarantees strictly equal-area pixels with a hierarchical multi-resolution structure.
Carlsson et al.~\cite{healswin} exploit this hierarchy for efficient spherical self-attention, while Krachmalnicoff et al.~\cite{healpix-convolutional} and Cheng et al.~\cite{cuhpx} provide foundational spherical CNN implementations.
Despite existing efforts for HEALPix-based applications~\cite{cuhpx}, HEALPix's irregular geometry has limitations in 3D tasks. 
We present the first CUDA implementation integrating HEALPix directly into the 3DGS rasterization pipeline.
\section{HEALPix-based 3D Gaussian Rasterization}
\label{sec:rasterization}

In this paper, we present a unified 3DGS framework that rasterizes directly on the HEALPix spherical grid. The following sections detail HEALPix adaptation for visual representation (\cref{sec:healpix}), latitude-aware spherical splatting with gradient derivations (\cref{sec:spherical_projection}), efficient tile query algorithms exploiting HEALPix's indexing (\cref{sec:tile_query}), and image synthesis for arbitrary camera models (\cref{sec:render}).

\subsection{Adapting HEALPix for Visual Representation}
\label{sec:healpix}

Our rasterizer operates on the HEALPix grid, which partitions the unit sphere into 12 base quadrilaterals, each recursively subdivided with resolution parameter $N_{\text{side}}$ (power of two), yielding $N_{\text{pix}} = 12 N_{\text{side}}^2$ equal-area pixels. 
This equal-area property ensures uniform solid angle $\Omega_{\mathrm{pix}} = 4\pi / N_{\mathrm{pix}}$ per pixel, providing balanced gradient contributions during backpropagation.

To dynamically match the input angular resolution, we compute the adaptive HEALPix resolution \(N_{\text{side}}^* = \sqrt{4\pi WH / (12\Omega_{\text{in}})}\) and round it to the nearest power of two. The solid angle \(\Omega_{\text{in}}\) is defined as \(4\pi\) for Equirectangular, \(2\pi(1 - \cos\theta)\) for Fisheye, and \(4\arcsin(\sin\theta_x \sin\theta_y)\) for Perspective, where \(\theta, \theta_x, \theta_y\) denote the respective half-FoVs.
For a camera model $\mathcal{C}$, let
$\mathcal{D}_{\mathcal{C}}\subseteq[0,W)\times[0,H)$ denote its valid image
domain. We define the image-to-sphere mapping
$\mathcal{P}_{\mathcal{C}}:
\mathcal{D}_{\mathcal{C}}\rightarrow\mathbb{S}^2$,
which maps each valid image pixel $(u,v)$ to its spherical direction
$(\omega,\phi)$. The corresponding visible spherical region is $\mathcal{R}_{\mathcal{C}} = \mathcal{P}_{\mathcal{C}}(\mathcal{D}_{\mathcal{C}})$, with the HEALPix visibility mask $ \mathcal{M}_{\mathcal{C}}(p) = \mathbf{1}\!\left[(\omega_p,\phi_p) \in\mathcal{R}_{\mathcal{C}}\right].$
Ground-truth images are resampled onto the adaptive-resolution HEALPix grid
within $\mathcal{R}_{\mathcal{C}}$, enabling unified supervision across camera
models, as shown in~\cref{fig:pipeline}(a).

HEALPix provides two indexing schemes: RING ordering arranges pixels along iso-latitude rings, while NESTED ordering follows the hierarchical quadtree structure, as illustrated in~\cref{fig:rasterization}.
We adopt NESTED indexing with $(x, y, f)$ parameterization, where $x, y \in [0, N_{\text{side}})$ are local coordinates within base quadrilateral $f \in [0, 12)$. This maps naturally to GPU parallelization: each base quadrilateral is partitioned into $\lceil N_{\text{side}}/B \rceil^2$ tiles of $B \times B$ threads, yielding $12 \lceil N_{\text{side}}/B \rceil^2$ parallel tiles with hierarchical spatial coherence for efficient memory access.

\subsection{Latitude-Aware Splatting on the Unit Sphere}
\label{sec:spherical_projection}

We now formulate how 3D Gaussians are splatted onto the HEALPix spherical screen space, parameterized in radians rather than pixels. Let $\mathbf{t} = \mathbf{W}\boldsymbol{\mu} + \mathbf{b}$ denote the camera-space position of a Gaussian center $\boldsymbol{\mu}$, where $\mathbf{W} \in \mathbb{R}^{3 \times 3}$ and $\mathbf{b}$ are the rotation and translation of the view matrix.

\textbf{Forward Projection.}
The spherical projection $\Pi: \mathbb{R}^3 \to \mathbb{S}^2$ maps $\mathbf{t} = (t_x, t_y, t_z)^\top$ to longitude-latitude coordinates:
\begin{equation}
\boldsymbol{\mu}' = \Pi(\mathbf{t}) = (\omega, \phi)^\top, \quad \omega = \mathrm{atan2}(t_x, t_z), \quad \phi = \mathrm{atan2}\left(t_y, \sqrt{t_x^2 + t_z^2}\right),
\label{eq:lonlat_projection}
\end{equation}
with radial depth $r = \|\mathbf{t}\|$ for depth sorting. The 3D covariance $\boldsymbol{\Sigma}_\mathrm{3D}$ is projected to radian-space covariance $\boldsymbol{\Sigma}_\mathrm{rad}$ via the Jacobian of $\Pi$ composed with the rotation, characterizing the Gaussian's angular extent for spherical rasterization.

Since the HEALPix screen measures arc length rather than angular displacement, and equal angular increments yield different arc lengths at different latitudes, we apply latitude-dependent scaling:
\begin{equation}
\boldsymbol{\Sigma}_\mathrm{arc} = \mathbf{S}_\phi \boldsymbol{\Sigma}_\mathrm{rad} \mathbf{S}_\phi^\top, \quad \mathbf{S}_\phi = \mathrm{diag}(\cos\phi, 1).
\label{eq:arc_scaling}
\end{equation}
Conic parameters are computed from $\boldsymbol{\Sigma}_\mathrm{arc}^{-1}$, and the spherical bounding radius $r_s = 3\sqrt{\lambda_{\max}(\boldsymbol{\Sigma}_\mathrm{arc})}$ determines tile overlap.

\textbf{Backward Propagation.}
Gradients propagate through this pipeline to update $\boldsymbol{\mu}$, scale $\mathbf{s}$, and rotation $\mathbf{q}$. Let $\mathcal{L}_X \coloneqq \frac{\partial \mathcal{L}}{\partial X}$.

\textbf{Covariance gradients.}
Since $\mathbf{S}_\phi$ depends on $\phi$, backpropagating through \cref{eq:arc_scaling} yields:
\begin{equation}
\mathcal{L}_{\boldsymbol{\Sigma}_\mathrm{rad}} = \mathbf{S}_\phi^\top \mathcal{L}_{\boldsymbol{\Sigma}_\mathrm{arc}} \mathbf{S}_\phi, \quad
\mathcal{L}_\phi\big|_{\mathbf{S}_\phi} = -\sin\phi \left( 2\cos\phi \, \sigma_{\omega\omega} \, \mathcal{L}_{\sigma_{\omega\omega}^\mathrm{arc}} + \sigma_{\omega\phi} \, \mathcal{L}_{\sigma_{\omega\phi}^\mathrm{arc}} \right),
\label{eq:backward_lat_scale}
\end{equation}
where $\boldsymbol{\Sigma}_\mathrm{rad} = \begin{psmallmatrix} \sigma_{\omega\omega} & \sigma_{\omega\phi} \\ \sigma_{\omega\phi} & \sigma_{\phi\phi} \end{psmallmatrix}$. The first term is the standard covariance gradient; the second captures how projected Gaussian shape varies with latitude. Gradients then propagate through $\boldsymbol{\Sigma}_\mathrm{rad} = \mathbf{T} \boldsymbol{\Sigma}_\mathrm{3D} \mathbf{T}^\top$ to update $\mathbf{s}$ and $\mathbf{q}$.

\textit{Position gradients.}
Beyond standard paths through $\mathbf{J}$ and $\boldsymbol{\mu}'$, spherical projection introduces an additional path through $\mathbf{S}_\phi$:
\begin{equation}
\frac{\partial \mathcal{L}}{\partial \mathbf{t}}\bigg|_{\mathbf{J}} = \frac{\partial \mathcal{L}}{\partial \boldsymbol{\Sigma}_\mathrm{rad}} \frac{\partial \boldsymbol{\Sigma}_\mathrm{rad}}{\partial \mathbf{J}} \frac{\partial \mathbf{J}}{\partial \mathbf{t}}, \quad
\frac{\partial \mathcal{L}}{\partial \mathbf{t}}\bigg|_{\mathbf{S}_\phi} = \frac{\partial \mathcal{L}}{\partial \boldsymbol{\Sigma}_\mathrm{arc}} \frac{\partial \boldsymbol{\Sigma}_\mathrm{arc}}{\partial \mathbf{S}_\phi} \frac{\partial \mathbf{S}_\phi}{\partial \phi} \frac{\partial \phi}{\partial \mathbf{t}}.
\label{eq:grad_via_J_S}
\end{equation}
The second term is a geometry-dependent gradient path unique to spherical projection. The total gradient, including $\frac{\partial \mathcal{L}}{\partial \boldsymbol{\mu}'} \mathbf{J}$, transforms to world space via $\frac{\partial \mathcal{L}}{\partial \boldsymbol{\mu}} = \mathbf{W}^\top \frac{\partial \mathcal{L}}{\partial \mathbf{t}}$. For depth supervision, we use inverse radial depth $r^{-1}$, whose gradient is $\frac{\partial r^{-1}}{\partial \mathbf{t}}=-\frac{\mathbf{t}}{r^3}$.

\subsection{Efficient Tile Query on HEALPix}
\label{sec:tile_query}

Given the projected spherical covariances, we perform tile-based culling to restrict each Gaussian's contribution to its intersecting HEALPix regions, ensuring high-performance rendering.
In vanilla 3DGS, each 2D Gaussian's rectangular footprint is intersected with a regular tile grid in $O(1)$ time. 
Since HEALPix pixels do not form a rectangular lattice, we develop two alternative query strategies exploiting its NESTED and RING indexing, which are illustrated in~\cref{fig:ablation}.

\textbf{NESTED Quadtree Traversal.}
HEALPix's NESTED indexing forms a quadtree hierarchy. 
Given a spherical disc with center $(\omega, \phi)$ and radius $r_s$, we perform depth-first search (DFS) from the 12 base quadrilaterals, pruning subtrees whose centers lie entirely outside the disc. 
Let $N_\mathrm{side}^{q}$ denote the query resolution, $P$ the number of Gaussians, and $K$ the average touched pixels per Gaussian.
This achieves $O(K + \log N_\mathrm{side}^{q})$ time per Gaussian, but requires $O(P \cdot \log N_\mathrm{side}^{q})$ stack memory and suffers from irregular memory access on the GPU.

\textbf{RING Sequential Scan.}
HEALPix's RING indexing arranges pixels in isolatitude rings, each characterized by $z = \cos\vartheta$ where $\vartheta \in [0, \pi]$ is the colatitude measured from the north pole. 
We compute the $z$-range intersecting the disc, then analytically solve for the longitude range within each ring:
\begin{equation}
\cos\Delta\omega = \frac{\cos r_s - z_c \cdot z_\mathrm{ring}}{\sqrt{1 - z_c^2} \cdot \sqrt{1 - z_\mathrm{ring}^2}},
\label{eq:ring_phi_range}
\end{equation}
where $z_c = \sin\phi_c$ is the $z$-coordinate of the disc center at latitude $\phi_c$, and $\Delta\omega$ is the half-width in longitude. 
This yields $O(K)$ time with $O(1)$ memory. 
Both schemes capture distinct trade-offs: NESTED prioritizes precision, while RING achieves a $1.7\times$ speedup at the expense of quality (see~\cref{sec:ablation}).

\subsection{Rendering on Spherical Grid and Image Synthesis}
\label{sec:render}

To ensure correct visibility across the full sphere, we perform per-tile alpha compositing by sorting Gaussians based on their radial distance $r = |\mathbf{t}|$ from the camera origin, rather than conventional planar $z$-depth.
Each HEALPix pixel center is determined by its NESTED index $(x, y, f)$, which maps to spherical coordinates $(\omega_\mathrm{pix}, \phi_\mathrm{pix})$ via the HEALPix indexing scheme.
During per-pixel blending, we compute the great-circle distance $d_\mathrm{gc}$ between each Gaussian center $(\omega, \phi)$ and the HEALPix pixel center using the Haversine formula~\cite{haversine}, as shown in~\cref{fig:pipeline} (b), then decompose it into local tangent-plane offsets $(d_x, d_y)$ via the spherical bearing angle. 
The per-pixel blending weight is computed as $\alpha = o \cdot \exp\bigl(-\tfrac{1}{2} \mathbf{d}^\top \boldsymbol{\Sigma}_\mathrm{arc}^{-1} \mathbf{d}\bigr)$, where $o$ is the Gaussian opacity and $\mathbf{d} = (d_x, d_y)^\top$. 

\begin{figure}[!t]
    \centering
    \includegraphics[width=\textwidth]{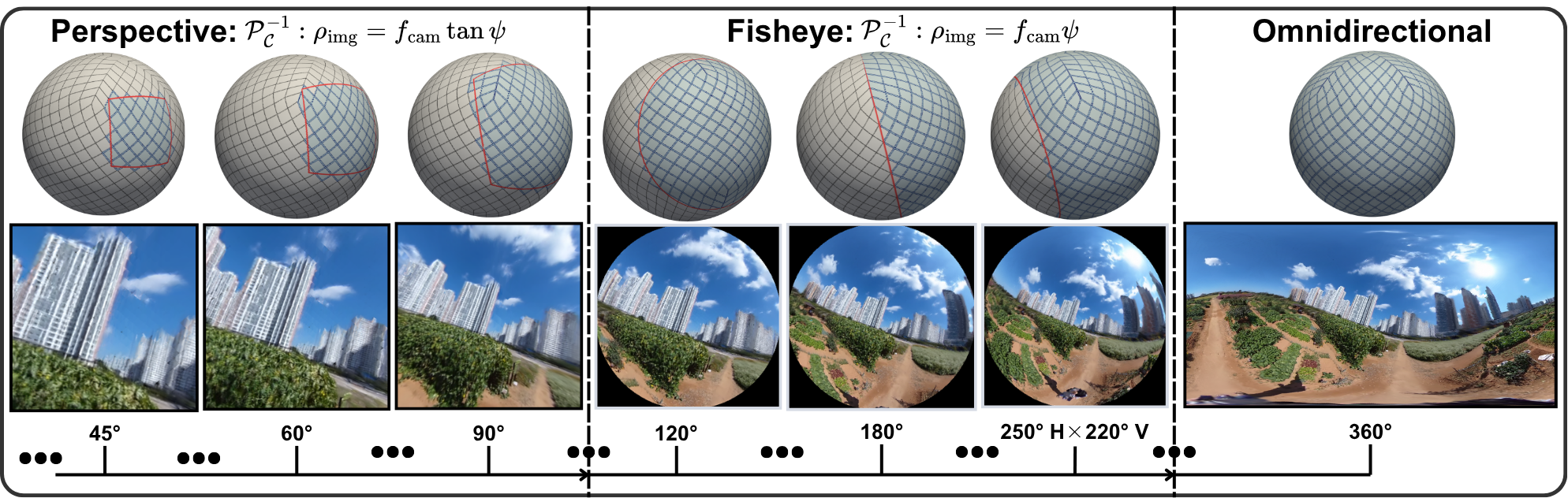}
    \caption{\textbf{Rendering across camera models and FoVs.} UniTriSplat synthesizes perspective, fisheye, and omnidirectional images by changing only the image-to-sphere mapping and valid FoV mask. The underlying spherical rasterization remains unchanged across camera configurations.}
    \label{fig:fov_support}
\end{figure}

Consequently, the HEALPix sphere is rendered through front-to-back
$\alpha$-blending. For a HEALPix pixel $p$, let $\mathcal{S}_p$ denote the set
of overlapping Gaussians, sorted by increasing radial depth $r_i$. Its rendered
color is
\begin{equation}
  I_{\mathrm{H}}(p)
  =
  \sum_{i\in\mathcal{S}_p}
  \mathbf{c}_i\,\alpha_i(p)
  \prod_{\substack{j\in\mathcal{S}_p\\r_j<r_i}}
  \left(1-\alpha_j(p)\right),
  \label{eq:unified_rendering}
\end{equation}
where $\mathbf{c}_i$ is the view-dependent color of Gaussian $i$, and
$\alpha_i(p)$ is the blending weight, evaluated for Gaussian $i$
at pixel $p$. Using the camera mapping $\mathcal{P}_{\mathcal{C}}$ and valid image domain
$\mathcal{D}_{\mathcal{C}}$ defined in~\cref{sec:healpix}, the final image is
obtained by interpolating neighboring HEALPix samples:
\begin{equation}
  I_{\mathcal{C}}(u,v)
  =
  \operatorname{Interp}_{\mathrm{H}}
  \!\left(I_{\mathrm{H}},\mathcal{P}_{\mathcal{C}}(u,v)\right),
  \qquad (u,v)\in\mathcal{D}_{\mathcal{C}}.
  \label{eq:camera_synthesis}
\end{equation}
For partial-FoV cameras, rendering and optimization are restricted to HEALPix
pixels satisfying $\mathcal{M}_{\mathcal{C}}(p)=1$, avoiding computation
outside the visible spherical region $\mathcal{R}_{\mathcal{C}}$. \Cref{fig:fov_support} illustrates image synthesis across different camera models and FoVs while keeping the underlying HEALPix rasterization unchanged.

\section{Training on the HEALPix Grid}
\label{sec:opt}

Having described the spherical rasterization pipeline, we now detail the training procedure. Two components require adaptation: the structure-aware loss must respect HEALPix's non-planar topology (\cref{sec:healssim}), and the density control thresholds must operate in radian space rather than pixel counts (\cref{sec:angular_density_control}).

\subsection{HEALPix-SSIM Loss}
\label{sec:healssim}

\begin{figure}[!t]
    \centering
    \includegraphics[width=1.0\textwidth]{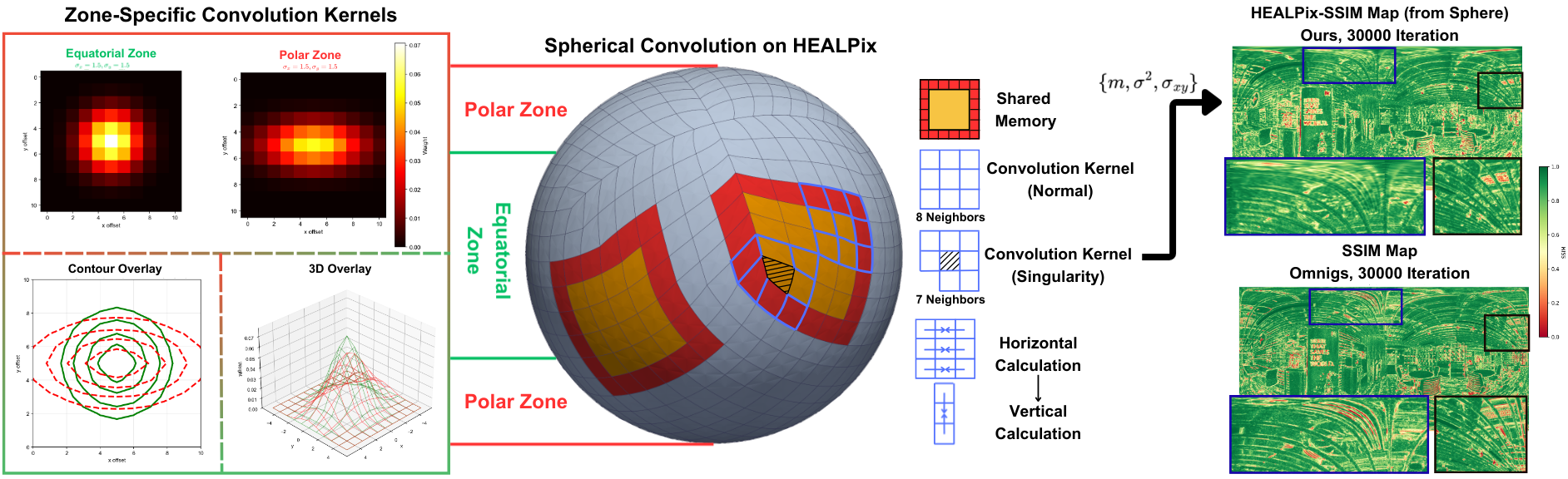} 
    \caption{\textbf{HEALPix-SSIM.} Zone-specific convolution kernels (left), spherical convolution with shared memory and singularity-aware boundary handling (middle), and the comparison of the resulting HSSIM map and SSIM map (right).}
    \label{fig:ssim} 
\end{figure}

The original 3DGS combines $\mathcal{L}_1$ photometric loss and SSIM~\cite{ssim} structure-aware loss. However, standard SSIM assumes a regular 2D grid, making it inapplicable to HEALPix tessellation. We propose HEALPix-SSIM (HSSIM), a structural similarity metric for the HEALPix grid with a fully differentiable CUDA implementation. Within each base quadrilateral, the NESTED $(x, y, f)$ parameterization approximates a local 2D grid, enabling separable convolution to compute local statistics $\mu$, $\sigma^2$, and covariance $\sigma_{12}$ between rendered and ground-truth HEALPix images for the SSIM index. At quadrilateral boundaries, topological singularities cause some corner pixels to have only 7 neighbors; we detect such cases and renormalize kernel weights accordingly. 
Furthermore, since equal pixel offsets correspond to different angular distances at varying latitudes, we employ zone-specific Gaussian kernels: isotropic for the equatorial zone where pixels are roughly square, and anisotropic for polar zones where pixel geometry is longitudinally compressed (\cref{fig:ssim}). Our CUDA implementation leverages shared memory with cross-face padding and caches intermediate derivatives for efficient gradient propagation via transposed convolution. The final training loss is $\mathcal{L} = (1 - \lambda) \mathcal{L}_1^\mathrm{H} + \lambda \, \mathcal{L}_\mathrm{H\text{-}SSIM}$, where $\mathcal{L}_1^\mathrm{H}$ is the per-pixel $\ell_1$ distance on the HEALPix grid and $\lambda$ balances photometric and perceptual terms.

\subsection{Radian-Space Density Control}
\label{sec:angular_density_control}

\begin{table*}[!t]
    \centering
    \caption{\textbf{Quantitative Results of Multi-FoV Evaluation.}
    We evaluate UniTriSplat on perspective (Pp.), fisheye (Fe.), and
    omnidirectional (Omni.) datasets. FoV denotes horizontal $\times$
    vertical field of view. Training time is reported for 30,000
    iterations. \colorbox{bestblue}{Blue} and
    \colorbox{secondblue}{light blue} indicate the best and second-best
    results, respectively.}
    \label{tab:multi_camera_evaluation}
    \small
    \setlength{\tabcolsep}{3pt}
    \begin{adjustbox}{max width=0.95\textwidth}
    \renewcommand{\arraystretch}{0.9}
    \begin{tabular}{@{}lllccccc}
        \toprule
        \textbf{Dataset} & \textbf{FoV} & \textbf{Method}
        & PSNR (dB)$\uparrow$ & SSIM$\uparrow$ & HSSIM$\uparrow$
        & LPIPS$\downarrow$ & Train (min)$\downarrow$ \\
        \midrule

        \multirow{3}{*}{\makecell[l]{\textbf{Pp.} \\ Mip-NeRF 360}}
        & \multirow{3}{*}{\makecell[c]{
        $52^\circ \times 36^\circ$--\\
        $68^\circ \times 47^\circ$}}
        & 3DGS             & 25.59 & 0.801 & 0.726 & 0.258 & \best{25.2} \\
        & & OP43DGS        & \best{28.03} & \best{0.879} & \second{0.741} & \best{0.231} & \second{45.6} \\
        & & \textbf{Ours}  & \second{27.57} & \second{0.848} & \best{0.806} & \second{0.239} & 47.2\\
        \cmidrule{1-8}

        \multirow{3}{*}{\makecell[l]{\textbf{Fe.} \\ ScanNet++}}
        & \multirow{3}{*}{$127^\circ \times 85^\circ$}
        & OP43DGS          & 21.82 & 0.728 & 0.650 & 0.355 & \best{30.5} \\
        & & Fisheye-GS     & \second{29.13} & \second{0.920} & \second{0.801} & \second{0.186} & \second{43.5} \\
        & & \textbf{Ours}  & \best{29.75} & \best{0.928} & \best{0.883} & \best{0.179} & 46.4 \\
        \cmidrule{1-8}

        \multirow{3}{*}{\makecell[l]{\textbf{Fe.} \\ FIORD}}
        & \multirow{3}{*}{\makecell[c]{
        $120^\circ \times 120^\circ$--\\
        $170^\circ \times 170^\circ$}}
        & OP43DGS          & \second{21.64} & \second{0.717} & \second{0.634} & \second{0.321} & \second{50.1} \\
        & & Fisheye-GS     & 21.15 & 0.704 & 0.615 & 0.350 & \best{25.6} \\
        & & \textbf{Ours}  & \best{24.27} & \best{0.791} & \best{0.748} & \best{0.279} & 54.2 \\
        \midrule

        \multirow{4}{*}{\makecell[l]{\textbf{Omni.} \\ Ricoh360}}
        & \multirow{4}{*}{$360^\circ \times 180^\circ$}
        & ODGS             & 21.89 & 0.813 & 0.786 & 0.246 & 42.8 \\
        & & OP43DGS        & 23.07 & 0.833 & 0.809 & 0.221 & 71.1 \\
        & & OmniGS         & \best{25.34} & \best{0.872} & \second{0.843} & \best{0.195} & \best{18.1} \\
        & & \textbf{Ours}  & \second{24.70} & \second{0.864} & \best{0.863} & \second{0.212} & \second{29.0} \\
        \cmidrule{1-8}

        \multirow{4}{*}{\makecell[l]{\textbf{Omni.} \\ OmniBlender}}
        & \multirow{4}{*}{$360^\circ \times 180^\circ$}
        & ODGS             & 29.81 & 0.876 & 0.859 & 0.239 & 40.5 \\
        & & OP43DGS        & 32.01 & 0.883 & 0.865 & 0.182 & 91.8 \\
        & & OmniGS         & \best{33.21} & \best{0.919} & \second{0.891} & \best{0.166} & \best{19.8} \\
        & & \textbf{Ours}  & \second{32.57} & \second{0.905} & \best{0.900} & \second{0.172} & \second{28.2} \\
        \cmidrule{1-8}

        \multirow{4}{*}{\makecell[l]{\textbf{Omni.} \\ 360Roam}}
        & \multirow{4}{*}{$360^\circ \times 180^\circ$}
        & ODGS             & 20.72 & 0.722 & 0.685 & 0.383 & 247.6 \\
        & & OP43DGS        & 21.04 & 0.732 & 0.686 & 0.376 & 330.7 \\
        & & OmniGS         & \second{21.48} & \second{0.741} & \second{0.710} & \second{0.369} & \second{122.6} \\
        & & \textbf{Ours}  & \best{21.82} & \best{0.747} & \best{0.724} & \best{0.353} & \best{102.1} \\

        \bottomrule
    \end{tabular}
    \end{adjustbox}
\end{table*}

Adaptive density control in 3DGS clones or splits high-gradient Gaussians and prunes low-opacity or oversized ones. 
Directly applying pixel-based thresholds to HEALPix is problematic because spherical projection maps Gaussians to angular coordinates $(\omega, \phi)$, requiring size-based criteria to operate in radian-space. 
For each visible Gaussian, the spherical projection produces a 2D covariance $\Sigma_\mathrm{arc}$ on the sphere, from which we derive an angular radius $r_s$ representing the Gaussian's spherical footprint. 
We maintain a per-Gaussian buffer that records the maximum $r_s$ observed across training iterations, updated via $r_{s,\max} \leftarrow \max(r_{s,\max}, r_s)$ whenever the Gaussian is visible. 
This radian-space tracking replaces pixel-based screen size and ensures scale comparisons remain consistent on the HEALPix grid. 
The pruning threshold for oversized Gaussians is similarly converted from pixels to radians based on the HEALPix resolution $N_\mathrm{side}$.

In order to reduce the number of primitives and time consumption, we also explore multi-view consistent pruning~\cite{fastgs}, which removes Gaussians contributing minimally to reconstruction across sampled views. This is an exploratory configuration that is not used in the final framework due to quality degradation. However, ablation results confirm that it provides a speed-quality trade-off.

\section{Experiment}

\begin{figure}[!t]
    \centering
    \includegraphics[width=1.0\textwidth]{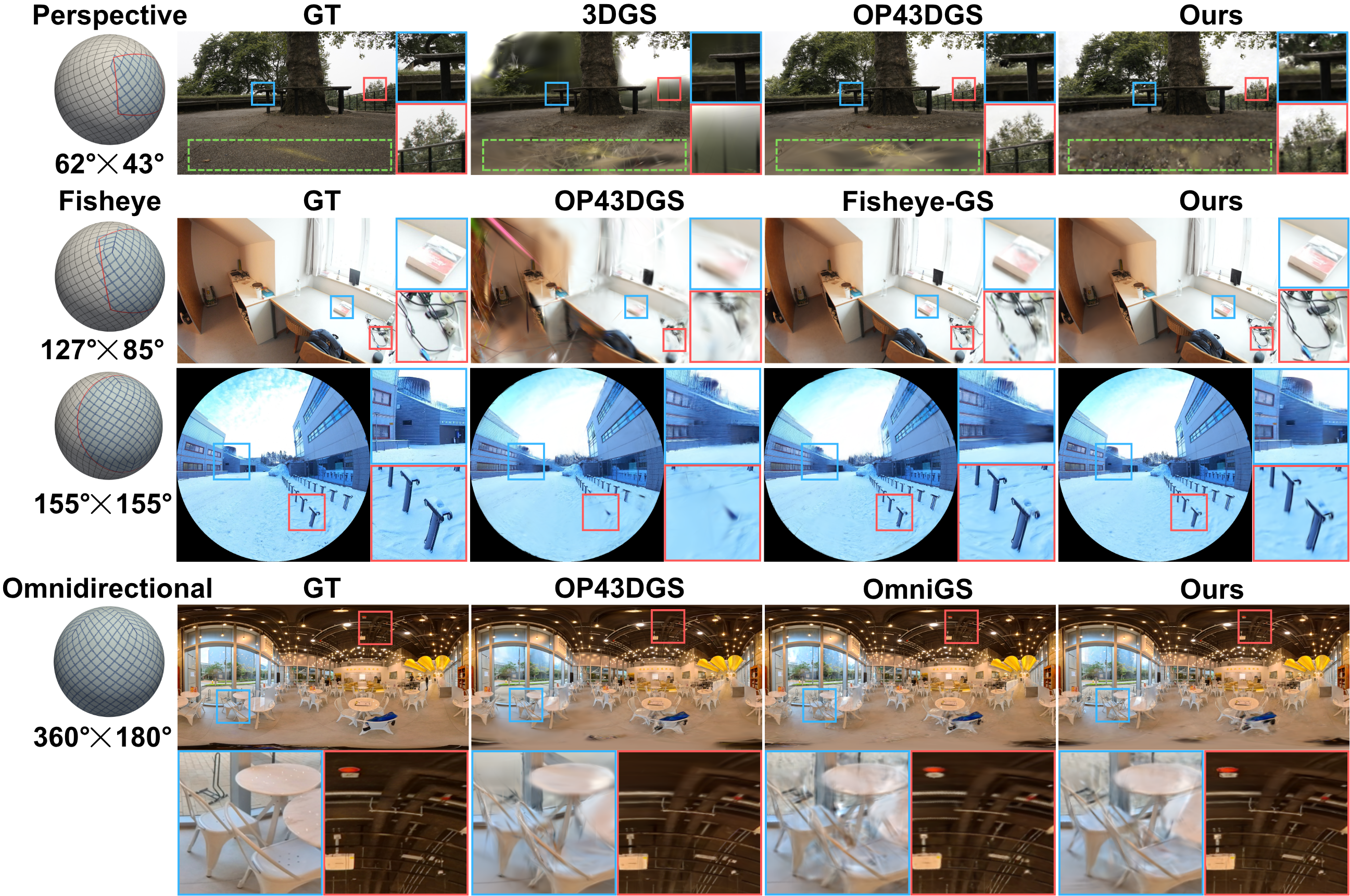} 
    \caption{\textbf{Qualitative Results of Multi-FoV Evaluation.} We compare UniTriSplat with camera-specific baselines across diverse camera models and FoVs. From top to bottom, the four rows show results on selected scenes from Mip-NeRF 360, ScanNet++, FIORD, and 360Roam, respectively. UniTriSplat maintains stable rendering quality, with better performance on fisheye and omnidirectional images. Additional results are provided in the supplementary material.} 
    \label{fig:exp-1} 
\end{figure}

In this section, we evaluate UniTriSplat's performance in terms of multi-FoV reconstruction fidelity, cross-camera generalization, and computational efficiency. We compare against 3DGS and OP43DGS~\cite{op43dgs} on perspective images, Fisheye-GS~\cite{fisheyegs} and
  OP43DGS on fisheye images, and ODGS~\cite{odgs}, OP43DGS, and OmniGS~\cite{omnigs} on omnidirectional images.
We conduct three sets of experiments: multi-FoV evaluation, cross-camera validation, and an ablation study.

\subsection{Implementation and Experiment Setup}
\label{sec:exp_settings}

\textbf{Implementation.}
UniTriSplat is built upon standard 3DGS with a redesigned rasterization pipeline and the HEALPix-aware SSIM module. While baseline methods retain their default training parameters, we adapt the densification thresholds, learning rates, and decay schedules for UniTriSplat to accommodate the unique gradient distributions inherent in the spherical domain (\cref{sec:spherical_projection}). 
We evaluate UniTriSplat using benchmarks covering multiple input FoV regimes: Mip-NeRF 360~\cite{mip} for perspective; ScanNet++~\cite{scannet++} and FIORD~\cite{fiord} for fisheye; and 360Roam~\cite{360roam}, OmniBlender~\cite{egonerf}, and Ricoh360~\cite{egonerf} for omnidirectional FoVs.
All models are trained on a single NVIDIA RTX 4090D GPU. 
 We report standard metrics including PSNR, SSIM, and LPIPS~\cite{unreasonable}. As traditional planar metrics can be biased against spherical representations due to resampling, we project baseline results onto the HEALPix grid to ensure a fair comparison via HEALPix-SSIM (HSSIM) (\cref{sec:healssim}).

\textbf{Multi-FoV evaluation.}
To assess UniTriSplat across diverse camera models, we evaluate it over a broad range of FoVs, including perspective, fisheye, and omnidirectional imaging. Each camera setting is evaluated on representative benchmarks against the corresponding projection-specific 3DGS baselines. This evaluates whether unified spherical rasterization maintains reconstruction quality across different projection distortion and spherical coverage.

\textbf{Cross-camera validation.}
To verify that the learned 3D Gaussians maintain consistency across diverse projections, we train on omnidirectional data and render perspective and fisheye views at multiple FoVs using our HEALPix rasterizer. We define the original camera orientation as the front direction. For Ricoh360, OmniBlender, and 360Roam, respectively, we render front-facing fisheye views with FoVs of $120^\circ$, $180^\circ$, and $240^\circ$, and perspective views with FoVs of $45^\circ$, $60^\circ$, and $90^\circ$ facing backward, rightward, and leftward. Ground-truth views are generated through sphere-to-plane sampling.

\subsection{Results, Evaluation and Discussion}

\begin{table*}[!t]
    \centering
    \caption{\textbf{Quantitative Results of Cross-camera Validation.}
    Given Gaussian scenes reconstructed from omnidirectional inputs, we evaluate
    perspective (Pp.) and fisheye (Fe.) rendering using camera-specific
    rasterizers and our unified rasterizer. \colorbox{bestblue}{Blue} and
    \colorbox{secondblue}{light blue} indicate the best and second-best results.
    Our method demonstrates consistent cross-camera generalization across
    diverse projection models.}
    \label{tab:omni_perspective}
    \begin{adjustbox}{max width=\textwidth}
        \begin{tabular}{ll|ccc|ccc|ccc|ccc}
            \toprule
            &
            & \multicolumn{3}{c|}{\textbf{Pp. from 3DGS}}
            & \multicolumn{3}{c|}{\textbf{Pp. from Ours}}
            & \multicolumn{3}{c|}{\textbf{Fe. from OP43DGS}}
            & \multicolumn{3}{c}{\textbf{Fe. from Ours}} \\
            \textbf{Dataset} & \textbf{Method}
            & PSNR$\uparrow$ & SSIM$\uparrow$ & LPIPS$\downarrow$
            & PSNR$\uparrow$ & SSIM$\uparrow$ & LPIPS$\downarrow$
            & PSNR$\uparrow$ & SSIM$\uparrow$ & LPIPS$\downarrow$
            & PSNR$\uparrow$ & SSIM$\uparrow$ & LPIPS$\downarrow$ \\
            \midrule

            \multirow{4}{*}{Ricoh360}
            & ODGS
            & 12.27 & 0.502 & 0.704
            & 17.59 & 0.602 & 0.511
            & 19.32 & 0.647 & 0.318
            & \second{22.52} & \second{0.707} & \second{0.294} \\
            & OP43DGS
            & 12.71 & 0.487 & 0.787
            & 18.81 & 0.693 & 0.440
            & \best{25.69} & \best{0.881} & \best{0.190}
            & 19.60 & 0.639 & 0.302 \\
            & OmniGS
            & \best{22.67} & \best{0.804} & \best{0.165}
            & \second{22.77} & \second{0.724} & \second{0.304}
            & \second{24.17} & \second{0.783} & \second{0.252}
            & 21.08 & 0.684 & 0.335 \\
            & \textbf{Ours}
            & \second{16.30} & \second{0.521} & \second{0.681}
            & \best{29.01} & \best{0.864} & \best{0.144}
            & 17.95 & 0.581 & 0.408
            & \best{27.66} & \best{0.893} & \best{0.147} \\
            \midrule

            \multirow{4}{*}{OmniBlender}
            & ODGS
            & 15.30 & 0.516 & 0.609
            & 20.40 & 0.623 & 0.379
            & 21.70 & 0.724 & 0.205
            & 22.98 & 0.811 & 0.315 \\
            & OP43DGS
            & 15.63 & 0.560 & 0.566
            & 20.76 & 0.646 & 0.363
            & \second{29.51} & \second{0.880} & \second{0.137}
            & 20.01 & 0.631 & 0.326 \\
            & OmniGS
            & \second{22.44} & \second{0.720} & \second{0.313}
            & \second{21.18} & \second{0.685} & \second{0.337}
            & 21.03 & 0.726 & 0.218
            & \second{23.65} & \second{0.832} & \second{0.293} \\
            & \textbf{Ours}
            & \best{22.54} & \best{0.759} & \best{0.262}
            & \best{28.17} & \best{0.834} & \best{0.168}
            & \best{31.33} & \best{0.905} & \best{0.128}
            & \best{31.94} & \best{0.916} & \best{0.134} \\
            \midrule

            \multirow{4}{*}{360Roam}
            & ODGS
            & 18.26 & 0.709 & 0.391
            & \second{23.65} & 0.759 & 0.333
            & 20.28 & 0.650 & 0.426
            & 20.74 & 0.683 & 0.404 \\
            & OP43DGS
            & 18.50 & 0.720 & 0.390
            & 23.61 & 0.766 & 0.326
            & 21.44 & 0.701 & 0.368
            & 21.09 & 0.689 & 0.385 \\
            & OmniGS
            & \second{18.75} & \second{0.731} & \second{0.389}
            & 23.55 & \second{0.773} & \second{0.319}
            & \second{22.03} & \second{0.729} & \second{0.327}
            & \second{22.40} & \second{0.741} & \second{0.299} \\
            & \textbf{Ours}
            & \best{22.02} & \best{0.765} & \best{0.333}
            & \best{24.70} & \best{0.786} & \best{0.304}
            & \best{23.87} & \best{0.809} & \best{0.276}
            & \best{23.99} & \best{0.815} & \best{0.261} \\

            \bottomrule
        \end{tabular}
    \end{adjustbox}
\end{table*}

\begin{figure}[!t]
    \centering
    \includegraphics[width=0.95\textwidth]{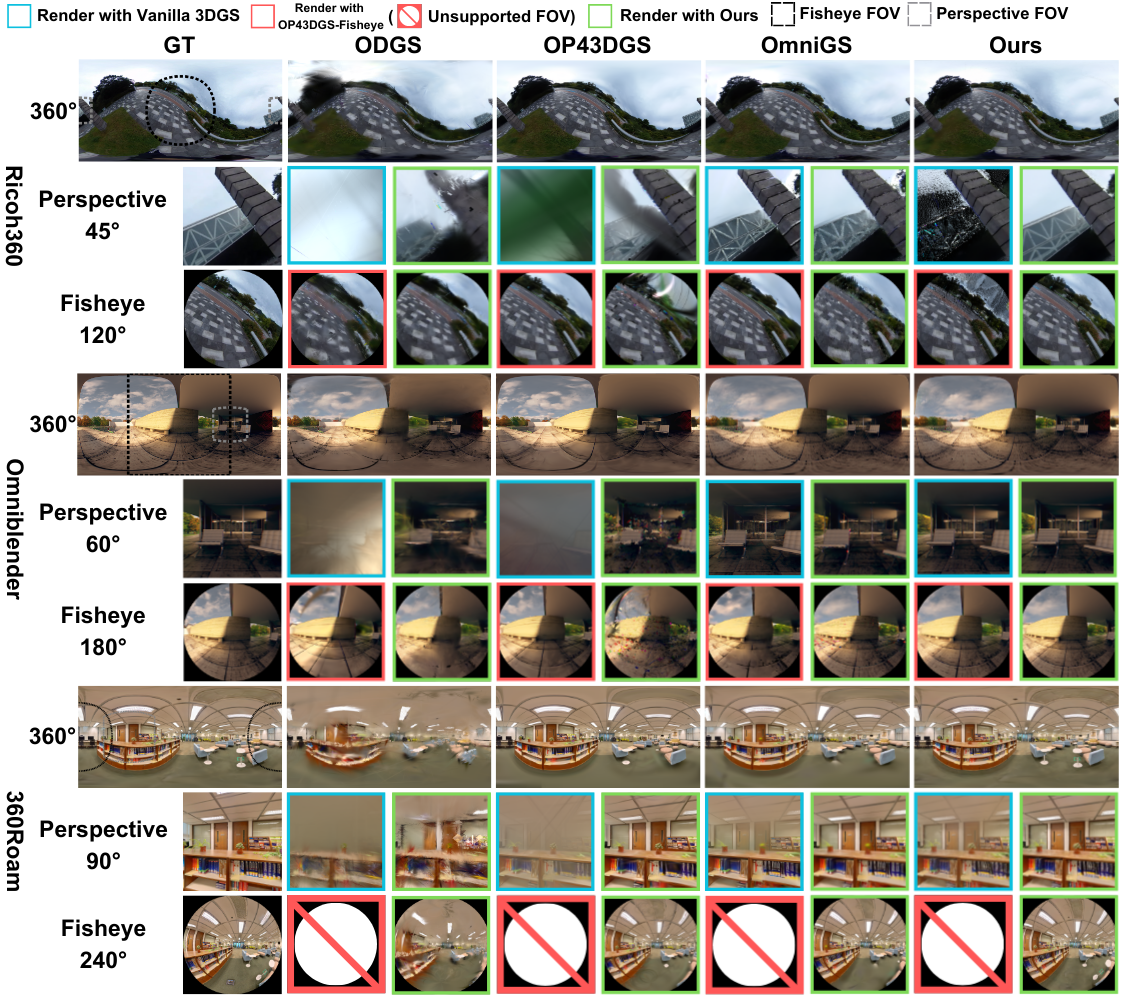} 
    \caption{\textbf{The Qualitative Results of Cross-camera Validation.} We evaluate renderings across different FoVs against GT images across three scenes (top to bottom: Pillar, Chair, Center). Our unified method maintains superior rendering quality and consistent geometric fidelity across heterogeneous cameras.} 
    \label{fig:exp-2} 
\end{figure}

\textbf{Multi-FoV Evaluation.} 
The quantitative results in~\cref{tab:multi_camera_evaluation} demonstrate that UniTriSplat maintains stable and competitive performance across multiple camera models and FoVs. On perspective inputs, UniTriSplat outperforms vanilla 3DGS~\cite{3dgs} and remains competitive with OP43DGS~\cite{op43dgs}. While spherical-to-planar resampling slightly reduces the sharpness, UniTriSplat achieves the best HSSIM, indicating improved structural consistency. In addition, UniTriSplat performs particularly well on severely distorted fisheye inputs. On the ultra-wide-FoV FIORD~\cite{fiord} dataset, UniTriSplat improves PSNR by approximately 3\,dB over the baselines, indicating that spherical-grid representations are inherently well suited to spherical imaging geometry. On ScanNet++~\cite{scannet++}, UniTriSplat also consistently improves all reconstruction metrics over Fisheye-GS~\cite{fisheyegs}, while the larger gain on FIORD suggests that spherical rasterization is effective under more severe distortion. On omnidirectional benchmarks, UniTriSplat achieves state-of-the-art performance on 360Roam~\cite{360roam} while remaining competitive on Ricoh360~\cite{egonerf} and OmniBlender~\cite{egonerf}. 
On the evaluated omnidirectional datasets, UniTriSplat maintains competitive training efficiency and records the shortest training time on 360Roam. Further efficiency analysis across input resolutions is provided in the supplementary material. Although OP43DGS supports multiple camera models, it is less competitive on the evaluated wide-FoV fisheye and omnidirectional datasets. This suggests that its camera-specific tangent-plane formulation is less robust under severe projection distortion. Moreover, UniTriSplat consistently achieves the best HSSIM across all datasets, demonstrating effective optimization over spherical geometry.

The qualitative results in~\cref{fig:exp-1} show that, by restricting HEALPix rasterization to the target FoV, UniTriSplat supports training and rendering across varying FoVs. On fisheye data, UniTriSplat produces fewer artifacts than the baselines, indicating that uniform spherical tessellation facilitates more accurate optimization under spherical projection. A similar effect is observed for omnidirectional images, where UniTriSplat better preserves polar details, such as the ceiling structures, which are degraded by the baselines. Although resampling from the sphere to the perspective plane slightly reduces sharpness, UniTriSplat produces cleaner boundaries at narrow FoVs, with fewer spike-shaped artifacts, and maintains consistent fidelity across the image.

\begin{figure}[!t]
    \centering
    \includegraphics[width=1.0\textwidth]{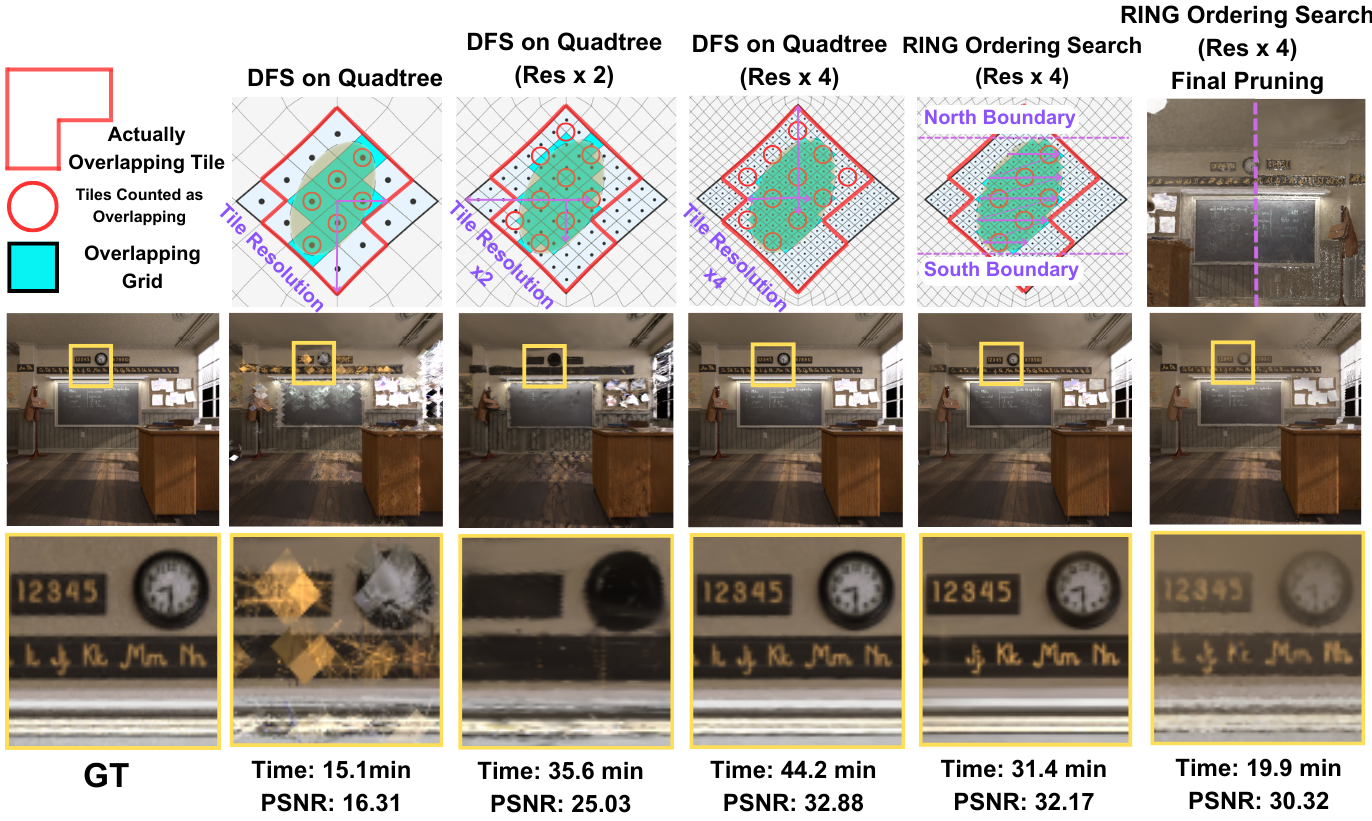} 
    \caption{\textbf{The illustration of Tile Query Methods and the Qualitative Results of the Ablation Study.} The `classroom' scene is reported here. DFS stands for depth-first search~\cite{introduction}. Comparison of NESTED quadtree traversal and RING sequential scan demonstrates quality-efficiency trade-offs across different query resolutions.}
    \label{fig:ablation} 
\end{figure}

\textbf{Cross-camera Validation.} 
\Cref{tab:omni_perspective} reveals a strong interaction between scene optimization and the rasterizer used for image synthesis. Camera-specific rasterizers generally perform best when applied to Gaussian representations optimized with their native projection model, but their performance can degrade substantially when transferred across camera models. In contrast, UniTriSplat maintains more consistent synthesis quality across perspective and fisheye projections, particularly for scenes optimized using its unified spherical rasterizer. These results suggest that HEALPix-based optimization reduces projection-specific coupling between the learned Gaussians and the camera model.

UniTriSplat generally achieves higher perspective-rendering quality than the vanilla 3DGS across the evaluated scenes. In fisheye synthesis, our rasterizer achieves competitive results across the evaluated scenes, except when OP43DGS renders the Gaussian representation optimized with its own rasterization pipeline. Notably, while OP43DGS is constrained by specific camera models and FoV limitations such as a sub-\(180^\circ\) range or the requirement for symmetric horizontal and vertical FoVs, UniTriSplat is camera-agnostic and handles arbitrary configurations. These results validate the effectiveness of our unified rasterization principle, demonstrating robust cross-camera generalization.

Qualitative results in~\cref{fig:exp-2} further substantiate these observations. Perspective crops synthesized by UniTriSplat exhibit higher fidelity with fewer geometric artifacts compared to vanilla 3DGS. 
A remaining challenge is observed in the Ricoh360 outdoor dataset, where OP43DGS rendering of our scenes shows artifacts in the sky regions, indicating a potential area for future optimization in handling unbounded backgrounds. Collectively, these results demonstrate that our unified spherical rasterization effectively decouples scene representation from camera-specific geometry, ensuring robust cross-camera generalization.

\subsection{Ablation Study}

We conduct ablation experiments on OmniBlender to analyze the contribution of each component in UniTriSplat. \Cref{tab:ablation} summarizes the results, where R denotes RING Sequential Scan, D denotes NESTED Quadtree Traversal (\cref{sec:tile_query}), and $\times 1$, $\times 2$, $\times 4$ indicate search depths at $1\times$, $2\times$, and $4\times$ the original tile resolution, respectively. The first four rows of the left column reveal that quadtree search on denser HEALPix grids improves reconstruction quality at the expense of time efficiency. RING Sequential Scan on denser grids achieves competitive metrics while maintaining high efficiency, and is therefore adopted as the default tile query module in UniTriSplat. As illustrated in~\cref{fig:ablation}, low-density deep search produces severe aliasing artifacts. Furthermore, removing either HSSIM or Density Control degrades performance across all reconstruction metrics, confirming the necessity of both modules. Final pruning substantially reduces the training time by removing redundant Gaussians. However, the resulting reduction in model capacity limits the achievable reconstruction quality, leading to lower metrics. We therefore exclude final pruning from our default configuration.

\label{sec:ablation}

\section{Conclusion}

\begin{table}[!t]
    \centering
    \caption{\textbf{Ablation study on the OmniBlender dataset.} We evaluate the impact of tile query strategies and optimization components. \textbf{R} denotes RING Sequential Scan, \textbf{D} denotes NESTED Quadtree Traversal (\cref{sec:tile_query}), and $\times n$ indicates the search depth relative to the tile resolution. The highlighted configuration \textbf{R ($\times 4$)} is selected as our default for its optimal performance-efficiency trade-off.}
    \resizebox{\columnwidth}{!}{
        \small
        \setlength{\tabcolsep}{2.5pt} 
        \begin{tabular}{llcccclccc}
            \toprule
            \multicolumn{2}{c}{\textbf{Tile Query}} & \textbf{PSNR}↑ & \textbf{SSIM}↑ & \textbf{Time (min)} & & \multicolumn{1}{c}{\textbf{Optimization}} & \textbf{PSNR}↑ & \textbf{SSIM}↑ & \textbf{Time (min)} \\
            \cmidrule(r){1-5} \cmidrule(l){7-10}
            \rowcolor{blue!10} 
            \multicolumn{2}{l}{R ($\times 4$)} & 32.57 & 0.905 & 28.2 & & Full Pipeline (Ref.) & 32.57 & 0.905 & 28.2 \\
            \multicolumn{2}{l}{D ($\times 4$)} & 33.64 & 0.916 & 49.1 & & w/o HSSIM & 26.42 & 0.721 & 21.7 \\
            \multicolumn{2}{l}{D ($\times 2$)} & 26.49 & 0.778 & 37.6 & & w/o Density Control & 24.83 & 0.732 & 33.0 \\
            \multicolumn{2}{l}{D ($\times 1$)} & 20.36 & 0.631 & 18.8 & & w/ Final Pruning & 28.27 & 0.847 & 20.2 \\
            \bottomrule
        \end{tabular}
    }
    \label{tab:ablation}
\end{table}

UniTriSplat introduces a unified 3DGS framework that reformulates rasterization on the equal-area HEALPix spherical tessellation, effectively mitigating the geometric distortions inherent in traditional planar projections. By decoupling spherical rasterization from camera-specific projection implementations, our approach enables model-agnostic synthesis for perspective, fisheye, and omnidirectional cameras within a single, consistent architecture. Leveraging a CUDA-accelerated rasterizer and a HEALPix-aware SSIM loss, UniTriSplat maintains stable reconstruction quality across diverse camera models and FoVs while enabling consistent cross-camera synthesis. Limitations include aliasing when resampling from HEALPix to planar images at lower resolutions, and minor projection errors from the EWA~\cite{ewa} splatting approximation. Future work will focus on refining rendering fidelity by addressing resampling aliasing and exploring advanced hierarchical acceleration structures tailored for the HEALPix topology.

\label{sec:conclusion}

\section*{Acknowledgements}
This work was partially supported by a grant from the Research Grants Council of the Hong Kong Special Administrative Region, China (Project No. HKUST 16202323), an internal grant from HKUST (R9429), the Frontier Technology Research for Joint Institutes with Industry (FTRIS) Project at HKUST (Project No. FTRIS-25-056), and an internal grant from Beijing Institute of Technology.
\bibliographystyle{splncs04}
\bibliography{main}

\begin{thebibliography}{10}
\providecommand{\url}[1]{\texttt{#1}}
\providecommand{\urlprefix}{URL }
\providecommand{\doi}[1]{https://doi.org/#1}

\bibitem{360gs}
Bai, J., Huang, L., Guo, J., Gong, W., Li, Y., Guo, Y.: 360-gs: Layout-guided
  panoramic gaussian splatting for indoor roaming. In: 2025 International
  Conference on 3D Vision (3DV). pp. 1042--1053. IEEE (2025)

\bibitem{mip}
Barron, J.T., Mildenhall, B., Verbin, D., Srinivasan, P.P., Hedman, P.:
  Mip-nerf 360: Unbounded anti-aliased neural radiance fields. In: Proceedings
  of the IEEE/CVF conference on computer vision and pattern recognition. pp.
  5470--5479 (2022)

\bibitem{healswin}
Carlsson, O., Gerken, J.E., Linander, H., Spie{\ss}, H., Ohlsson, F.,
  Petersson, C., Persson, D.: Heal-swin: A vision transformer on the sphere.
  In: Proceedings of the IEEE/CVF Conference on Computer Vision and Pattern
  Recognition. pp. 6067--6077 (2024)

\bibitem{quicktime}
Chen, S.E.: Quicktime vr: An image-based approach to virtual environment
  navigation. In: Proceedings of the 22nd annual conference on Computer
  graphics and interactive techniques. pp. 29--38 (1995)

\bibitem{cuhpx}
Cheng, X., Subramaniam, A., Wu, S., Brenowitz, N.: cuhpx: Gpu-accelerated
  differentiable spherical harmonic transforms on healpix grids. arXiv preprint
  arXiv:2510.01785  (2025)

\bibitem{sphericaltransformer}
Cho, S., Jung, R., Kwon, J.: Spherical transformer. arXiv preprint
  arXiv:2202.04942  (2022)

\bibitem{egonerf}
Choi, C., Kim, S.M., Kim, Y.M.: Balanced spherical grid for egocentric view
  synthesis. In: Proceedings of the IEEE/CVF Conference on Computer Vision and
  Pattern Recognition. pp. 16590--16599 (2023)

\bibitem{spherenet}
Coors, B., Condurache, A.P., Geiger, A.: Spherenet: Learning spherical
  representations for detection and classification in omnidirectional images.
  In: Proceedings of the European conference on computer vision (ECCV). pp.
  518--533 (2018)

\bibitem{introduction}
Cormen, T.H., Leiserson, C.E., Rivest, R.L., Stein, C.: Introduction to
  algorithms. MIT press (2022)

\bibitem{deng2025self}
Deng, Y., Xian, W., Yang, G., Guibas, L., Wetzstein, G., Marschner, S.,
  Debevec, P.: Self-calibrating gaussian splatting for large field-of-view
  reconstruction. In: Proceedings of the IEEE/CVF International Conference on
  Computer Vision. pp. 25124--25133 (2025)

\bibitem{healpix}
Gorski, K.M., Hivon, E., Banday, A.J., Wandelt, B.D., Hansen, F.K., Reinecke,
  M., Bartelmann, M.: Healpix: A framework for high-resolution discretization
  and fast analysis of data distributed on the sphere. The Astrophysical
  Journal  \textbf{622}(2),  759--771 (2005)

\bibitem{environment}
Greene, N.: Environment mapping and other applications of world projections.
  IEEE computer graphics and Applications  \textbf{6}(11),  21--29 (1986)

\bibitem{fiord}
Gunes, U., Turkulainen, M., Ren, X., Solin, A., Kannala, J., Rahtu, E.: Fiord:
  A fisheye indoor-outdoor dataset with lidar ground truth for 3d scene
  reconstruction and benchmarking. In: Scandinavian Conference on Image
  Analysis. pp. 3--17. Springer (2025)

\bibitem{360dvo}
Guo, X., Xu, Y., Huang, H., Yeung, S.K.: 360dvo: Deep visual odometry for
  monocular 360-degree camera. IEEE Robotics and Automation Letters
  \textbf{11}(3),  3079--3086 (2026)

\bibitem{scomnigs}
Huang, H., Chen, Y., Li, L., Cheng, H., Braud, T., Zhao, Y., Yeung, S.K.:
  Sc-omnigs: Self-calibrating omnidirectional gaussian splatting. In: The
  Thirteenth International Conference on Learning Representations (2025)

\bibitem{360roam}
Huang, H., Chen, Y., Zhang, T., Yeung, S.K.: 360roam: Real-time indoor roaming
  using geometry-aware 360$^\circ$ radiance fields. arXiv preprint
  arXiv:2208.02705  (2022)

\bibitem{360loc}
Huang, H., Liu, C., Zhu, Y., Cheng, H., Braud, T., Yeung, S.K.: 360loc: A
  dataset and benchmark for omnidirectional visual localization with
  cross-device queries. In: Proceedings of the IEEE/CVF conference on computer
  vision and pattern recognition. pp. 22314--22324 (2024)

\bibitem{360vot}
Huang, H., Xu, Y., Chen, Y., Yeung, S.K.: 360vot: A new benchmark dataset for
  omnidirectional visual object tracking. In: Proceedings of the IEEE/CVF
  International Conference on Computer Vision. pp. 20566--20576 (2023)

\bibitem{op43dgs}
Huang, L., Bai, J., Guo, J., Li, Y., Guo, Y.: On the error analysis of 3d
  gaussian splatting and an optimal projection strategy. In: European
  conference on computer vision. pp. 247--263. Springer (2024)

\bibitem{erpgs}
Ito, S., Takama, N., Ito, K., Chen, H.T., Aoki, T.: Erpgs: Equirectangular
  image rendering enhanced with 3d gaussian regularization. In: 2025 IEEE
  International Conference on Image Processing (ICIP). pp. 2850--2855. IEEE
  (2025)

\bibitem{vrgs}
Jiang, Y., Yu, C., Xie, T., Li, X., Feng, Y., Wang, H., Li, M., Lau, H., Gao,
  F., Yang, Y., Jiang, C.: Vr-gs: A physical dynamics-aware interactive
  gaussian splatting system in virtual reality. arXiv preprint arXiv:2401.16663
   (2024)

\bibitem{yinyang}
Kageyama, A., Sato, T.: “yin-yang grid”: An overset grid in spherical
  geometry. Geochemistry, Geophysics, Geosystems  \textbf{5}(9) (2004)

\bibitem{3dgs}
Kerbl, B., Kopanas, G., Leimk{\"u}hler, T., Drettakis, G., et~al.: 3d gaussian
  splatting for real-time radiance field rendering. ACM Trans. Graph.
  \textbf{42}(4),  139--1 (2023)

\bibitem{healpix-convolutional}
Krachmalnicoff, N., Tomasi, M.: Convolutional neural networks on the healpix
  sphere: a pixel-based algorithm and its application to cmb data analysis.
  Astronomy \& Astrophysics  \textbf{628}, ~A129 (2019)

\bibitem{odgs}
Lee, S., Chung, J., Huh, J., Lee, K.M.: Odgs: 3d scene reconstruction from
  omnidirectional images with 3d gaussian splattings. Advances in Neural
  Information Processing Systems  \textbf{37},  57050--57075 (2024)

\bibitem{omnisplat}
Lee, S., Chung, J., Kim, K., Huh, J., Lee, G., Lee, M., Lee, K.M.: Omnisplat:
  Taming feed-forward 3d gaussian splatting for omnidirectional images with
  editable capabilities. In: Proceedings of the Computer Vision and Pattern
  Recognition Conference. pp. 16356--16365 (2025)

\bibitem{spags}
Li, J., Hahlbohm, F., Scholz, T., Eisemann, M., Tauscher, J.P., Magnor, M.:
  Spags: Fast and accurate 3d gaussian splatting for spherical panoramas. In:
  Computer Graphics Forum. vol.~44, p. e70171. Wiley Online Library (2025)

\bibitem{omnigs}
Li, L., Huang, H., Yeung, S.K., Cheng, H.: Omnigs: Fast radiance field
  reconstruction using omnidirectional gaussian splatting. In: 2025 IEEE/CVF
  Winter Conference on Applications of Computer Vision (WACV). pp. 2260--2268.
  IEEE (2025)

\bibitem{fisheyegs}
Liao, Z., Chen, S., Fu, R., Wang, Y., Su, Z., Luo, H., Ma, L., Xu, L., Dai, B.,
  Li, H., et~al.: Fisheye-gs: Lightweight and extensible gaussian splatting
  module for fisheye cameras. arXiv preprint arXiv:2409.04751  (2024)

\bibitem{fastgs}
Ren, S., Wen, T., Fang, Y., Lu, B.: Fastgs: Training 3d gaussian splatting in
  100 seconds. arXiv preprint arXiv:2511.04283  (2025)

\bibitem{icosahedral-hexagonal}
Sadourny, R., Arakawa, A., Mintz, Y.: Integration of the nondivergent
  barotropic vorticity equation with an icosahedral-hexagonal grid for the
  sphere. Monthly Weather Review  \textbf{96}(6),  351--356 (1968)

\bibitem{seam360gs}
Shin, C., Cho, W.O., Kim, S.J.: Seam360gs: Seamless 360deg gaussian splatting
  from real-world omnidirectional images. In: Proceedings of the IEEE/CVF
  International Conference on Computer Vision. pp. 28970--28979 (2025)

\bibitem{mapping}
Snyder, J.P.: Flattening the earth: two thousand years of map projections.
  University of Chicago Press (1997)

\bibitem{vrsplat}
Tu, X., Radl, L., Steiner, M., Steinberger, M., Kerbl, B., De~la Torre, F.:
  Vrsplat: Fast and robust gaussian splatting for virtual reality. Proceedings
  of the ACM on Computer Graphics and Interactive Techniques  \textbf{8}(1),
  1--22 (2025)

\bibitem{haversine}
Van~Brummelen, G.: Heavenly mathematics: The forgotten art of spherical
  trigonometry. Princeton University Press (2012)

\bibitem{ssim}
Wang, Z., Bovik, A.C., Sheikh, H.R., Simoncelli, E.P.: Image quality
  assessment: from error visibility to structural similarity. IEEE transactions
  on image processing  \textbf{13}(4),  600--612 (2004)

\bibitem{3dgut}
Wu, Q., Esturo, J.M., Mirzaei, A., Moenne-Loccoz, N., Gojcic, Z.: 3dgut:
  Enabling distorted cameras and secondary rays in gaussian splatting. In:
  Proceedings of the Computer Vision and Pattern Recognition Conference. pp.
  26036--26046 (2025)

\bibitem{street}
Yan, Y., Lin, H., Zhou, C., Wang, W., Sun, H., Zhan, K., Lang, X., Zhou, X.,
  Peng, S.: Street gaussians: Modeling dynamic urban scenes with gaussian
  splatting. In: ECCV (2024)

\bibitem{scannet++}
Yeshwanth, C., Liu, Y.C., Nie{\ss}ner, M., Dai, A.: Scannet++: A high-fidelity
  dataset of 3d indoor scenes. In: Proceedings of the IEEE/CVF International
  Conference on Computer Vision. pp. 12--22 (2023)

\bibitem{unreasonable}
Zhang, R., Isola, P., Efros, A.A., Shechtman, E., Wang, O.: The unreasonable
  effectiveness of deep features as a perceptual metric. In: Proceedings of the
  IEEE conference on computer vision and pattern recognition. pp. 586--595
  (2018)

\bibitem{distortioncnn}
Zhao, Q., Zhu, C., Dai, F., Ma, Y., Jin, G., Zhang, Y.: Distortion-aware cnns
  for spherical images. In: IJCAI. pp. 1198--1204 (2018)

\bibitem{hugs}
Zhou, H., Shao, J., Xu, L., Bai, D., Qiu, W., Liu, B., Wang, Y., Geiger, A.,
  Liao, Y.: Hugs: Holistic urban 3d scene understanding via gaussian splatting.
  In: Proceedings of the IEEE/CVF Conference on Computer Vision and Pattern
  Recognition (CVPR). pp. 21336--21345 (June 2024)

\bibitem{ewa}
Zwicker, M., Pfister, H., Van~Baar, J., Gross, M.: Ewa volume splatting. In:
  Proceedings Visualization, 2001. VIS'01. pp. 29--538. IEEE (2001)

\end{thebibliography}


\begin{thebibliography}{1}
\providecommand{\url}[1]{\texttt{#1}}
\providecommand{\urlprefix}{URL }
\providecommand{\doi}[1]{https://doi.org/#1}

\bibitem{mip}
Barron, J.T., Mildenhall, B., Verbin, D., Srinivasan, P.P., Hedman, P.:
  Mip-nerf 360: Unbounded anti-aliased neural radiance fields. In: Proceedings
  of the IEEE/CVF conference on computer vision and pattern recognition. pp.
  5470--5479 (2022)

\bibitem{fiord}
Gunes, U., Turkulainen, M., Ren, X., Solin, A., Kannala, J., Rahtu, E.: Fiord:
  A fisheye indoor-outdoor dataset with lidar ground truth for 3d scene
  reconstruction and benchmarking. In: Scandinavian Conference on Image
  Analysis. pp. 3--17. Springer (2025)

\bibitem{scannet++}
Yeshwanth, C., Liu, Y.C., Nie{\ss}ner, M., Dai, A.: Scannet++: A high-fidelity
  dataset of 3d indoor scenes. In: Proceedings of the IEEE/CVF International
  Conference on Computer Vision. pp. 12--22 (2023)

\end{thebibliography}

\newpage
{
    \centering
    \bfseries\Large UniTriSplat: A Unified 3D Gaussian Splatting Framework with Uniform Spherical Rasterization for Universal Cameras\\[10pt]
    \bfseries\large --- Supplementary Material ---
    \par
}

\suppressfloats[t]

\section{Supplementary Experimental Details}
\label{sec:supp_exp_details}

This section provides additional details of the datasets and evaluation protocols used in the multi-FoV experiments. We first summarize the benchmark coverage, then describe the common HEALPix-based computation of HSSIM and the treatment of stitching artifacts in omnidirectional images.

\subsection{Datasets and Evaluation Coverage}
\label{sec:supp_datasets}

The selected datasets cover perspective, fisheye, and omnidirectional imaging, with FoVs ranging from conventional perspective views to complete omnidirectional coverage. They also combine real and synthetic scenes, indoor and outdoor environments, and different image resolutions. This diversity allows us to evaluate whether the same spherical rasterization framework remains effective under changes in camera geometry, scene content, and acquisition conditions. Mip-NeRF 360 contains both bounded indoor scenes and unbounded outdoor scenes. ScanNet++ provides calibrated indoor fisheye images, whereas FIORD contains challenging ultra-wide FoV indoor and outdoor scenes. Ricoh360, 360Roam, and OmniBlender complement these datasets with real outdoor omnidirectional scenes, high-resolution real indoor omnidirectional scenes, and synthetic omnidirectional scenes, respectively. \Cref{tab:supp_datasets} shows the summary of the datasets we use in the multi-FoV evaluation.

\subsection{Fair HSSIM Evaluation}
\label{sec:supp_hssim_fairness}

HSSIM serves two distinct roles in our framework: it is used as a spherical structural loss for UniTriSplat and as a common evaluation metric for all methods. Although optimizing the HSSIM loss improves UniTriSplat under this metric, the baselines similarly optimize standard SSIM in their native image domains. Moreover, the variant without the HSSIM loss still improves standard SSIM over the baselines on many scenes, indicating that the reconstruction gains do not solely result from optimizing the evaluation metric. HSSIM is included to measure structural fidelity on uniformly sampled spherical geometry, where planar SSIM is affected by projection-dependent pixel density.

For a camera model $\mathcal{C}$, let $\Pi_{\mathcal{C}}^{-1}:\mathbb{S}^2\rightarrow\mathbb{R}^2$ map a spherical direction to its image coordinate. Each HEALPix pixel $q$ corresponds to a unit direction $\mathbf{s}_q\in\mathbb{S}^2$. Given a rendered image $I_{\mathcal{C}}^{m}$ from method $m$ and its ground truth $G_{\mathcal{C}}$, we construct their HEALPix representations using the same projection and interpolation:
\begin{equation}
    I_{\mathrm{H}}^{m}(q)
    = I_{\mathcal{C}}^{m}\!\left(\Pi_{\mathcal{C}}^{-1}(\mathbf{s}_q)\right),
    \qquad
    G_{\mathrm{H}}(q)
    = G_{\mathcal{C}}\!\left(\Pi_{\mathcal{C}}^{-1}(\mathbf{s}_q)\right).
    \label{eq:supp_image_to_healpix}
\end{equation}
Only directions that project into the valid camera domain are retained through a visibility mask
\begin{equation}
    M_{\mathcal{C}}(q)
    = \mathbf{1}\!\left[\Pi_{\mathcal{C}}^{-1}(\mathbf{s}_q)\in\mathcal{D}_{\mathcal{C}}\right],
    \label{eq:supp_hssim_mask}
\end{equation}
where $\mathcal{D}_{\mathcal{C}}$ is the valid image region determined by the camera model and FoV. The local HEALPix-SSIM value $h_q$ is computed over the spherical neighborhood $\mathcal{N}_{\mathrm{H}}(q)$ using the boundary-aware kernels described in the main paper. The final score is the masked average
\begin{equation}
    \operatorname{HSSIM}(I_{\mathcal{C}}^{m},G_{\mathcal{C}})
    = \frac{\sum_q M_{\mathcal{C}}(q)\,h_q\!\left(I_{\mathrm{H}}^{m},G_{\mathrm{H}}\right)}
    {\sum_q M_{\mathcal{C}}(q)}.
    \label{eq:supp_hssim_eval}
\end{equation}
All baseline and UniTriSplat outputs therefore undergo the same camera-to-sphere mapping, visibility masking, interpolation, and HSSIM computation. This protocol isolates structural quality on the sphere from the nonuniform sampling of the original image projection.

\subsection{Preprocessing of Omnidirectional Images}
\label{sec:supp_stitching}

\begin{table*}[!t]
    \centering
    \caption{\textbf{Datasets used in the multi-FoV evaluation.} Resolution is reported as width $\times$ height after preprocessing. FoV denotes horizontal $\times$ vertical field of view. The scene count refers to the subset used in our evaluation.}
    \label{tab:supp_datasets}
    \small
    \setlength{\tabcolsep}{4pt}
    \begin{adjustbox}{max width=\textwidth}
    \begin{tabular}{@{}llllcll}
        \toprule
        \textbf{Camera} & \textbf{Dataset} & \textbf{Resolution} & \textbf{FoV} & \textbf{Scenes} & \textbf{Environment} & \textbf{Source} \\
        \midrule
        Perspective
        & Mip-NeRF 360~\citesupp{mip}
        & $1237\times822$
        & \makecell[l]{$52^\circ\times36^\circ$--\\$68^\circ\times47^\circ$}
        & 9 & Indoor/Outdoor & Real \\
        \midrule
        \multirow{2}{*}{Fisheye}
        & ScanNet++~\citesupp{scannet++}
        & $1752\times1168$
        & $127^\circ\times85^\circ$
        & 7 & Indoor & Real \\
        & FIORD~\citesupp{fiord}
        & $3264\times3264$
        & \makecell[l]{$120^\circ\times120^\circ$--\\$170^\circ\times170^\circ$}
        & 7 & Indoor/Outdoor & Real \\
        \midrule
        \multirow{3}{*}{Omnidirectional}
        & Ricoh360~\cite{egonerf}
        & $1920\times960$
        & $360^\circ\times180^\circ$
        & 12 & Outdoor & Real \\
        & 360Roam~\cite{360roam}
        & $6080\times3040$
        & $360^\circ\times180^\circ$
        & 11 & Indoor & Real \\
        & OmniBlender~\cite{egonerf}
        & $2000\times1000$
        & $360^\circ\times180^\circ$
        & 8 & Indoor/Outdoor & Synthetic \\
        \bottomrule
    \end{tabular}
    \end{adjustbox}
\end{table*}

Commercial $360^\circ$ cameras commonly form an omnidirectional image by calibrating, warping, and blending images from multiple fisheye lenses. This upstream imaging process may introduce seam discontinuities, exposure differences, or local geometric misalignment. For datasets that provide processed omnidirectional images, we use the released stitched frames directly. For raw dual-fisheye captures, standard preprocessing first maps both lenses to a common spherical coordinate system using the calibrated camera parameters, applies photometric correction in overlapping regions, and blends the overlap before conversion to an equirectangular image. The resulting omnidirectional images are used consistently for pose estimation, training, and evaluation by all methods.

UniTriSplat does not contain a dedicated seam-removal module. Stitching artifacts arise from the physical multi-lens capture and preprocessing pipeline rather than the 3DGS rasterization model. Handling such hardware-specific artifacts is therefore outside the main scope of this work. Our objective is to provide a unified rasterizer for the processed perspective, fisheye, and omnidirectional images. More advanced calibration and seam-aware preprocessing can be incorporated independently in future work.

\section{Additional Evaluation}
\label{sec:supp_additional_evaluation}

We supplement the multi-FoV results with a controlled cross-resolution evaluation on 360Roam. This experiment analyzes how the discrete HEALPix hierarchy affects reconstruction quality, runtime, and resource consumption as the input resolution changes.

\begin{table*}[!t]
    \centering
    \caption{\textbf{Cross-resolution evaluation on 360Roam.} Ours(H) and Ours(L) use the HEALPix subdivision levels immediately above and below the target input pixel count, respectively. Blue and light blue denote the best and second-best results within each resolution.}
    \label{tab:supp_cross_resolution}
    \scriptsize
    \setlength{\tabcolsep}{2.2pt}
    \renewcommand{\arraystretch}{1.05}
    \begin{adjustbox}{max width=\textwidth}
    \begin{tabular}{@{}clccccccccc}
        \toprule
        \multirow{2}{*}{\textbf{Resolution}} & \multirow{2}{*}{\textbf{Method}}
        & \multicolumn{4}{c}{\textbf{Reconstruction}}
        & \multicolumn{2}{c}{\textbf{Runtime}}
        & \multicolumn{3}{c}{\textbf{Resources}} \\
        \cmidrule(lr){3-6}\cmidrule(lr){7-8}\cmidrule(lr){9-11}
        & & PSNR$\uparrow$ & SSIM$\uparrow$ & LPIPS$\downarrow$ & HSSIM$\uparrow$
        & \makecell{Train\\(min)$\downarrow$} & \makecell{Render\\(ms)$\downarrow$}
        & \makecell{Static GPU\\Mem. (MB)$\downarrow$} & \makecell{Peak GPU\\Mem. (MB)$\downarrow$}
        & \makecell{Avg.\\\#Gaussians$\downarrow$} \\
        \midrule
        \multirow{5}{*}{$6080\times3040$}
        & ODGS            & 20.72 & 0.722 & 0.383 & 0.685 & 247.6 & 70.44 & 1740.2 & 17176.6 & 598872 \\
        & OP43DGS         & 21.04 & 0.732 & 0.376 & 0.686 & 330.7 & 83.21 & 1867.9 & 15560.9 & 683659 \\
        & OmniGS          & 21.48 & 0.741 & 0.369 & 0.710 & \second{122.6} & \best{23.65} & \best{848.4} & \best{5983.3} & \second{465993} \\
        & \textbf{Ours(H)} & \best{22.53} & \best{0.759} & \best{0.346} & \best{0.733} & 324.4 & 92.01 & 1216.4 & 8926.4 & 521740 \\
        & \textbf{Ours(L)} & \second{21.82} & \second{0.747} & \second{0.353} & \second{0.724} & \best{102.1} & \second{40.80} & \second{1035.5} & \second{7037.7} & \best{454153} \\
        \midrule
        \multirow{5}{*}{$3040\times1520$}
        & ODGS            & 22.43 & 0.736 & 0.353 & 0.701 & 67.1 & 32.79 & 1092.1 & 10482.3 & 1025384 \\
        & OP43DGS         & 23.10 & 0.783 & 0.305 & 0.737 & 150.0 & 51.22 & 842.3 & 6158.2 & 779652 \\
        & OmniGS          & 23.61 & 0.793 & 0.299 & 0.752 & \best{33.5} & \best{15.33} & \best{582.0} & \best{3724.8} & \second{768071} \\
        & \textbf{Ours(H)} & \best{24.37} & \best{0.811} & \best{0.276} & \best{0.795} & 97.8 & 38.25 & 1176.4 & 9826.1 & 932404 \\
        & \textbf{Ours(L)} & \second{23.99} & \second{0.798} & \second{0.283} & \second{0.778} & \second{50.4} & \second{30.96} & \second{713.8} & \second{4812.5} & \best{705181} \\
        \midrule
        \multirow{5}{*}{$1520\times760$}
        & ODGS            & 22.93 & 0.758 & 0.254 & 0.711 & 30.3 & 19.67 & 826.9 & 6217.1 & 1342318 \\
        & OP43DGS         & 23.51 & 0.774 & 0.230 & 0.756 & 65.1 & 31.87 & 635.2 & 3924.7 & 1063944 \\
        & OmniGS          & \second{25.05} & \second{0.809} & \second{0.187} & 0.763 & \best{20.6} & \second{11.09} & \best{438.6} & \best{2478.5} & \best{791826} \\
        & \textbf{Ours(H)} & \best{26.10} & \best{0.827} & \best{0.167} & \best{0.821} & 46.7 & 25.50 & 895.5 & 5843.6 & 1101735 \\
        & \textbf{Ours(L)} & 23.84 & 0.796 & 0.198 & \second{0.775} & \second{25.2} & \best{10.46} & \second{541.0} & \second{3186.9} & \second{943072} \\
        \bottomrule
    \end{tabular}
    \end{adjustbox}
\end{table*}

\subsection{Cross-Resolution Protocol}
\label{sec:supp_cross_resolution_protocol}

We construct three image scales from the high-resolution 360Roam dataset: $6080\times3040$, $3040\times1520$, and $1520\times760$. All methods use the same train/test split, calibrated camera poses, SfM initialization, 30,000 optimization iterations, and a single NVIDIA GeForce RTX 4090 D GPU. The baseline methods retain their official optimization settings, while UniTriSplat uses the spherical-domain learning-rate and densification scaling described in~\cref{sec:hyperpara}. The final-pruning variant is not used in any reported result.

HEALPix supports only discrete subdivision levels. We therefore evaluate two adjacent configurations. \textbf{Ours(H)} selects the HEALPix level immediately above the target input pixel count, while \textbf{Ours(L)} selects the level immediately below it. Ours(H) prioritizes sampling density and reconstruction quality; Ours(L) reduces the number of sampled spherical pixels and targets a better efficiency--quality trade-off. Training time measures the complete 30,000-iteration optimization, and rendering time is averaged over the test views. Static GPU memory is measured after model initialization, whereas peak memory records the maximum allocated GPU memory during training. The number of Gaussians is averaged over the evaluated scenes after training.

\begin{figure}[!t]
    \centering
    \includegraphics[width=1.0\textwidth]{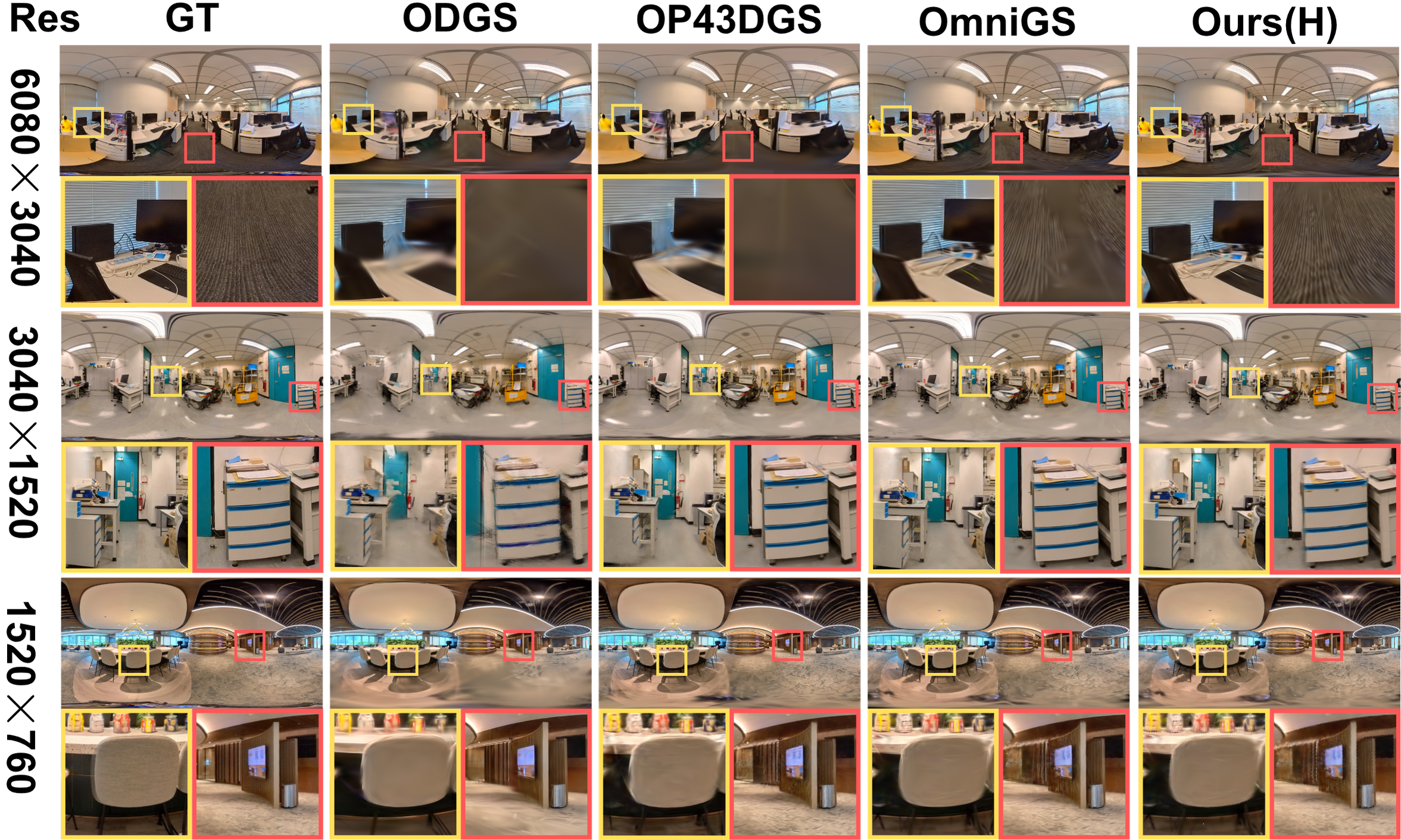}
    \caption{\textbf{Qualitative cross-resolution comparison.} From top to bottom, the Office, Lab, and Cafe scenes from 360Roam are evaluated at decreasing input resolutions. The enlarged regions emphasize texture and polar details. Ours(H) preserves fine structures more consistently, whereas the lower HEALPix level may lose high-frequency details after sphere-to-image resampling.}
    \label{fig:supp_cross_resolution}
\end{figure}

\subsection{Cross-Resolution Results}
\label{sec:supp_cross_resolution_results}

As shown in~\cref{tab:supp_cross_resolution}, Ours(H) achieves the best reconstruction metrics at all three resolutions, confirming that a denser HEALPix level increases representation capacity. Ours(L) provides a different operating point: it remains competitive at medium and high resolutions while reducing the number of sampled spherical pixels. At $6080\times3040$, Ours(L) completes training in 102.1 minutes, compared with 122.6 minutes for OmniGS, and uses the fewest Gaussians among all methods. Its static and peak GPU memory are higher than OmniGS but substantially lower than ODGS and OP43DGS.

The efficiency gain of Ours(L) at the highest resolution is associated with approximately $31.9\%$ fewer sampled pixels than the matched equirectangular grid, which offsets the CUDA overhead of HEALPix indexing and sphere-to-image sampling. This benefit is resolution-dependent rather than universal. At lower resolutions, the fixed indexing and resampling costs account for a larger fraction of runtime, and the lower HEALPix level may lose high-frequency information during sphere-to-image sampling. This explains why Ours(L) is less accurate than OmniGS at $1520\times760$, whereas Ours(H) recovers the missing detail by using the next subdivision level. Ours(L) nevertheless attains the lowest rendering latency at this scale.

\begin{figure}[!t]
    \centering
    \includegraphics[width=0.8\textwidth]{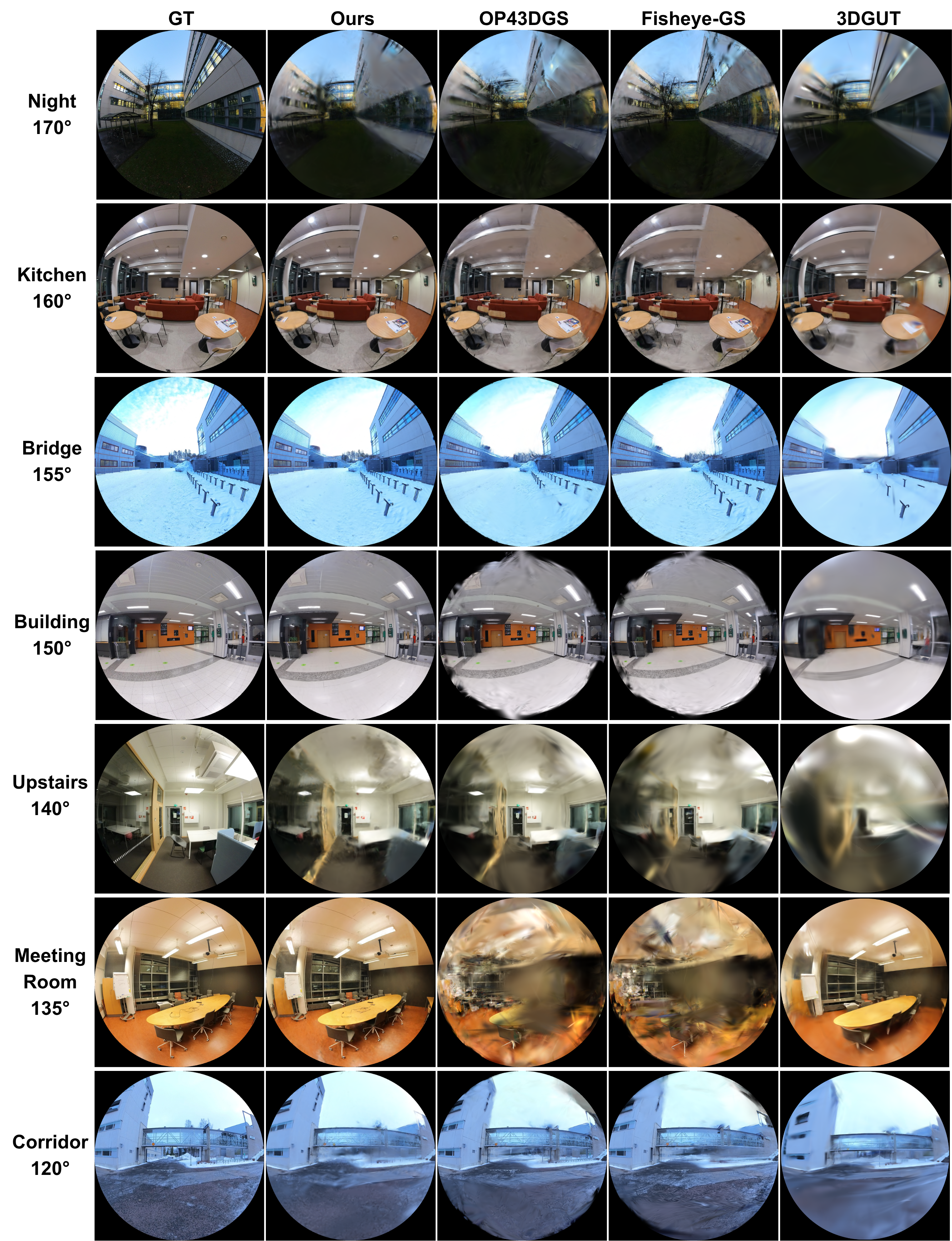}
    \caption{Qualitative comparison on FIORD using fisheye training inputs. UniTriSplat is compared with OP43DGS, Fisheye-GS, and 3DGUT.}
    \label{fig:supp_exp2}
\end{figure}

The qualitative comparisons in~\cref{fig:supp_cross_resolution} support the quantitative results. In the high-resolution Office scene, Ours(H) preserves carpet texture that is smoothed by the baseline. In the Cafe scene, it retains fine ceiling structures near the viewing poles even after downsampling. These observations are consistent with the equal-area sampling of HEALPix, which avoids the strong polar oversampling of equirectangular images. At the same time, the softer output of Ours(L) at low resolution exposes the remaining limitation of sphere-to-image interpolation.

\begin{figure}[!t]
    \centering
    \includegraphics[width=0.95\textwidth]{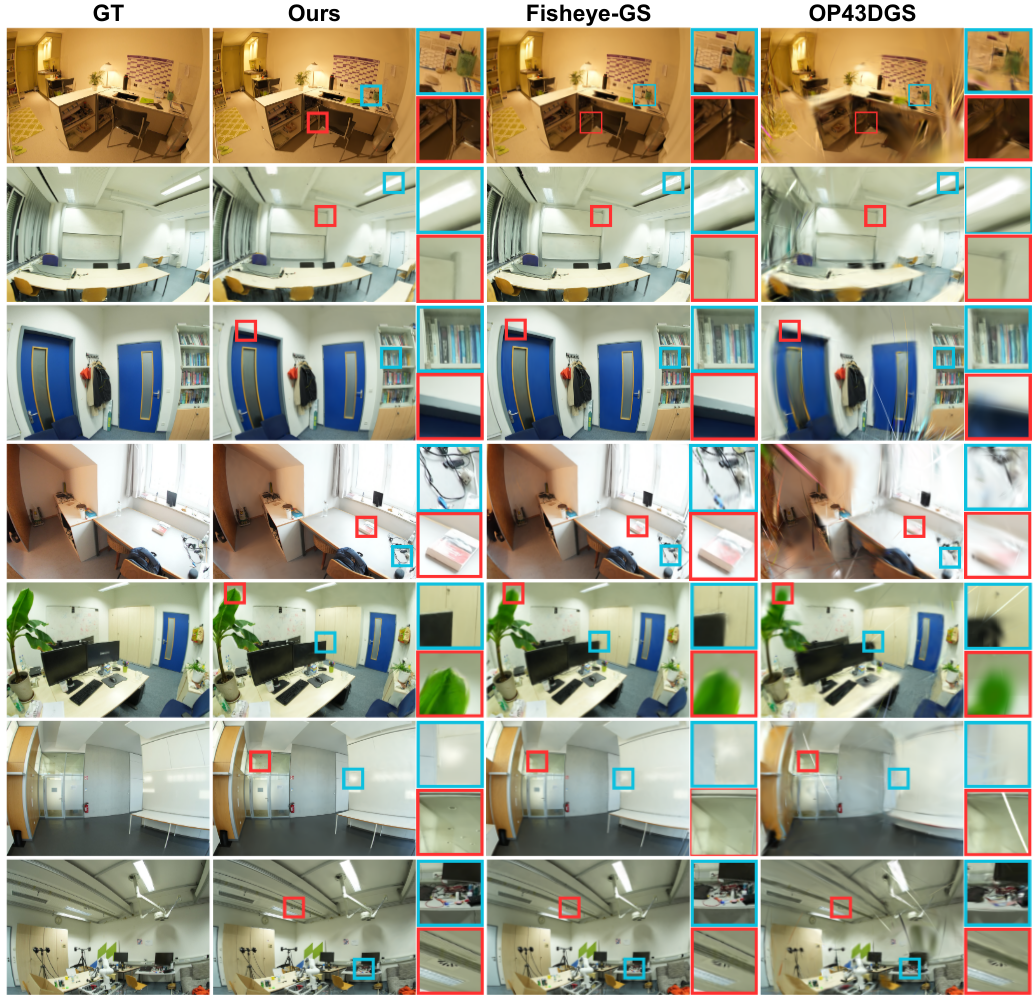}
    \caption{Qualitative comparison on ScanNet++ using fisheye training inputs. UniTriSplat is compared with Fisheye-GS.}
    \label{fig:supp_exp3}
\end{figure}

\subsection{Efficiency Limitations and Future Work}
\label{sec:supp_efficiency_limitations}

The generality of spherical rasterization introduces overhead that is absent from camera-specific planar pipelines. HEALPix indexing, spherical tile queries, and the final sphere-to-image sampling require additional CUDA operations. These costs are amortized for high-resolution omnidirectional inputs, where equal-area sampling removes substantial polar redundancy, but they are more visible for low-resolution or narrow-FoV images. Future work will simplify HEALPix indexing and search, improve sphere-to-planar reconstruction filters, and develop specialized low-level GPU kernels. These optimizations target lower overhead without changing the unified camera-model formulation.

\begin{figure}[!t]
    \centering
    \includegraphics[width=1.0\textwidth]{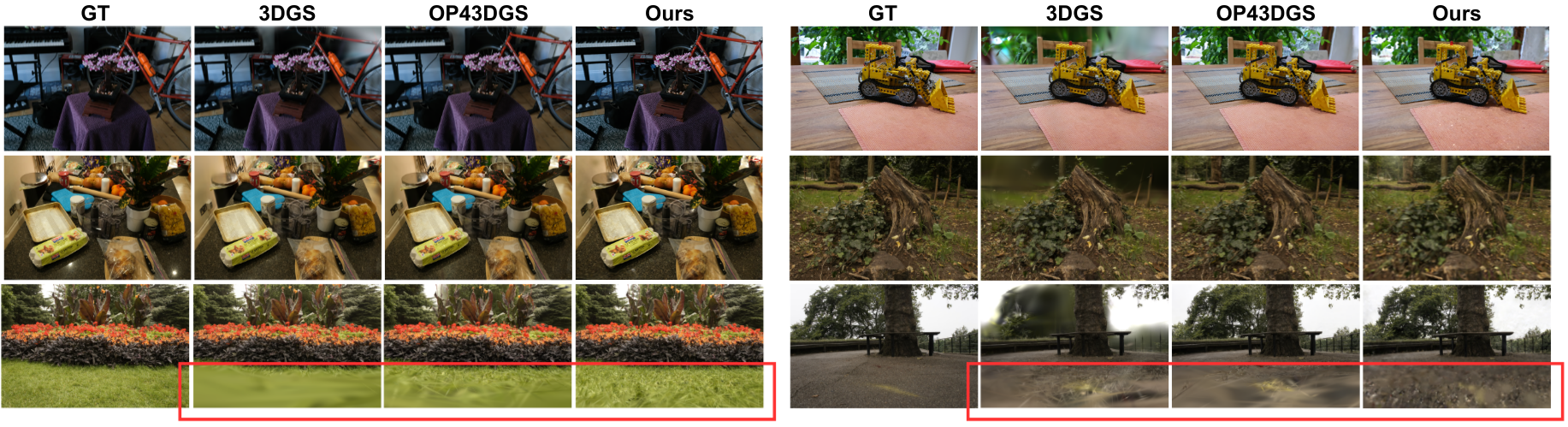}
    \caption{Qualitative comparison on Mip-NeRF 360 using perspective training inputs. UniTriSplat is compared with vanilla 3DGS and OP43DGS using the pinhole model.}
    \label{fig:supp_exp4}
\end{figure}

\section{Additional Qualitative Results}
\label{sec:supp_qualitative}

This section provides additional qualitative comparisons for fisheye and perspective training inputs and for cross-camera rendering of Gaussian scenes reconstructed by vanilla 3DGS.

\subsection{Fisheye Image Inputs}
\label{sec:eva-fisheye}

For fisheye training, we compare UniTriSplat with Fisheye-GS and OP43DGS on two datasets. Moreover, we also compare with 3DGUT on the FIORD dataset. Fisheye-GS and OP43DGS incorporate fisheye projection into EWA splatting, whereas UniTriSplat optimizes the scene on the common HEALPix sphere. We use the equidistant fisheye model for all methods. For datasets calibrated with polynomial distortion models, the input images are undistorted and reprojected using the provided calibration before training. We also modify the OP43DGS fisheye rasterizer to accept explicit camera intrinsics rather than deriving focal length only from a symmetric FoV.

As shown in~\cref{fig:supp_exp2,fig:supp_exp3}, UniTriSplat produces fewer floating artifacts and better preserves local structures under strong fisheye distortion. The difference is most visible on FIORD, whose ultra-wide FoV covers regions close to and beyond the hemisphere boundary. ScanNet++ contains a smaller FoV and more accurate pose registration; on these indoor scenes, the improvements are concentrated around fine structures and image boundaries. The FIORD Night scene remains more challenging because low illumination and severe distortion reduce feature-matching accuracy and degrade the SfM poses.

\subsection{Perspective Image Inputs}
\label{sec:eva-perspective}

For perspective training, we compare UniTriSplat with vanilla 3DGS and OP43DGS on Mip-NeRF 360. The spherical representation provides stable Gaussian footprints near image boundaries, but the final resampling from the HEALPix sphere to a perspective image can reduce sharpness. This effect is more visible for narrow-FoV images because they have a higher angular resolution than panoramic inputs at the same image resolution.

In~\cref{fig:supp_exp4}, UniTriSplat recovers scene geometry consistently and produces fewer thin, spike-shaped artifacts near image boundaries. In contrast, its output is slightly softer than vanilla 3DGS and OP43DGS in some high-frequency regions. The equal-area spherical representation avoids the strong stretching of projected Gaussian footprints near perspective-image boundaries, while the remaining blur results from interpolation between spherical and planar samples. Improving this resampling step is an important direction for future work.

\subsection{Cross-Camera Validation on Vanilla 3DGS Models}
\label{sec:cross-cam-vanilla}

We further evaluate cross-camera rendering on Gaussian scenes reconstructed by vanilla 3DGS from perspective images. Without retraining the Gaussian representation, we use different rasterizers to synthesize fisheye and omnidirectional views. This experiment separates the compatibility of the rasterizer from the optimization procedure used to obtain the Gaussian scene.

\begin{figure}[!t]
    \centering
    \includegraphics[width=0.6\textwidth]{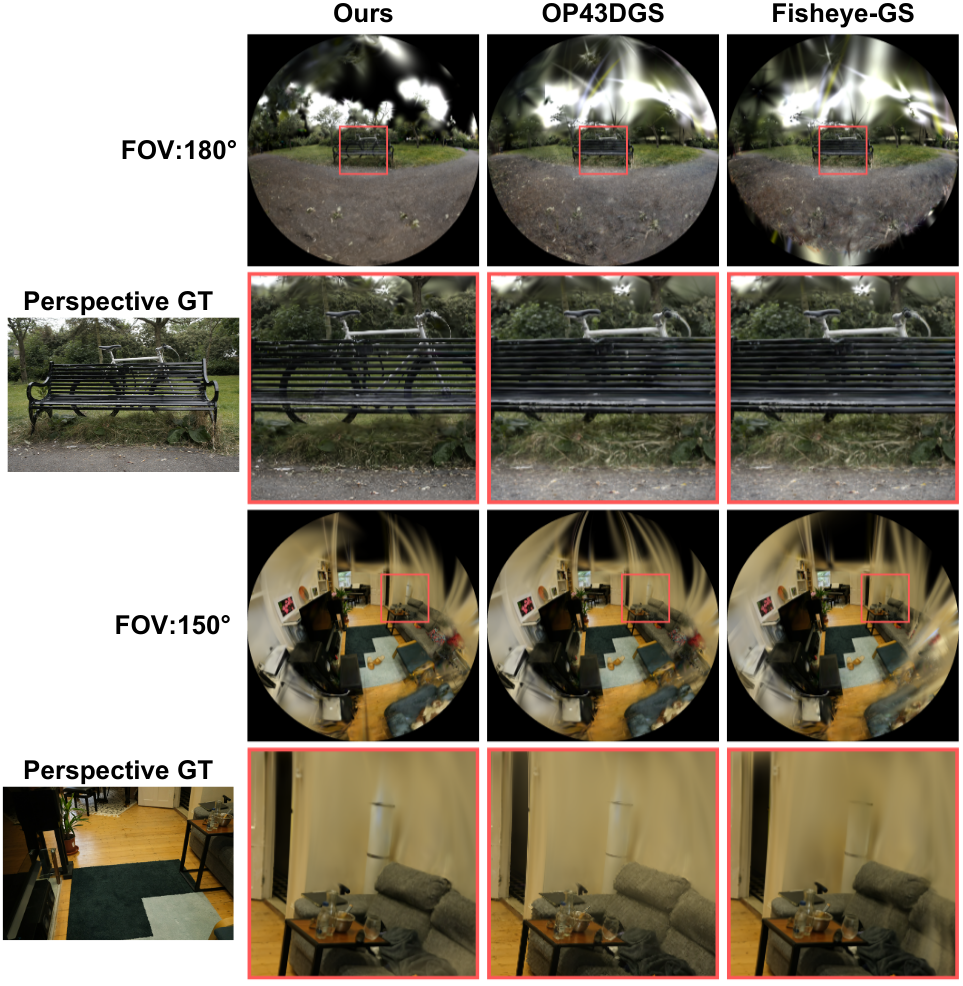}
    \caption{Fisheye rendering of vanilla-3DGS models trained on Mip-NeRF 360, compared with OP43DGS and Fisheye-GS.}
    \label{fig:supp_exp1}
\end{figure}

\begin{figure}[!t]
    \centering
    \includegraphics[width=0.7\textwidth]{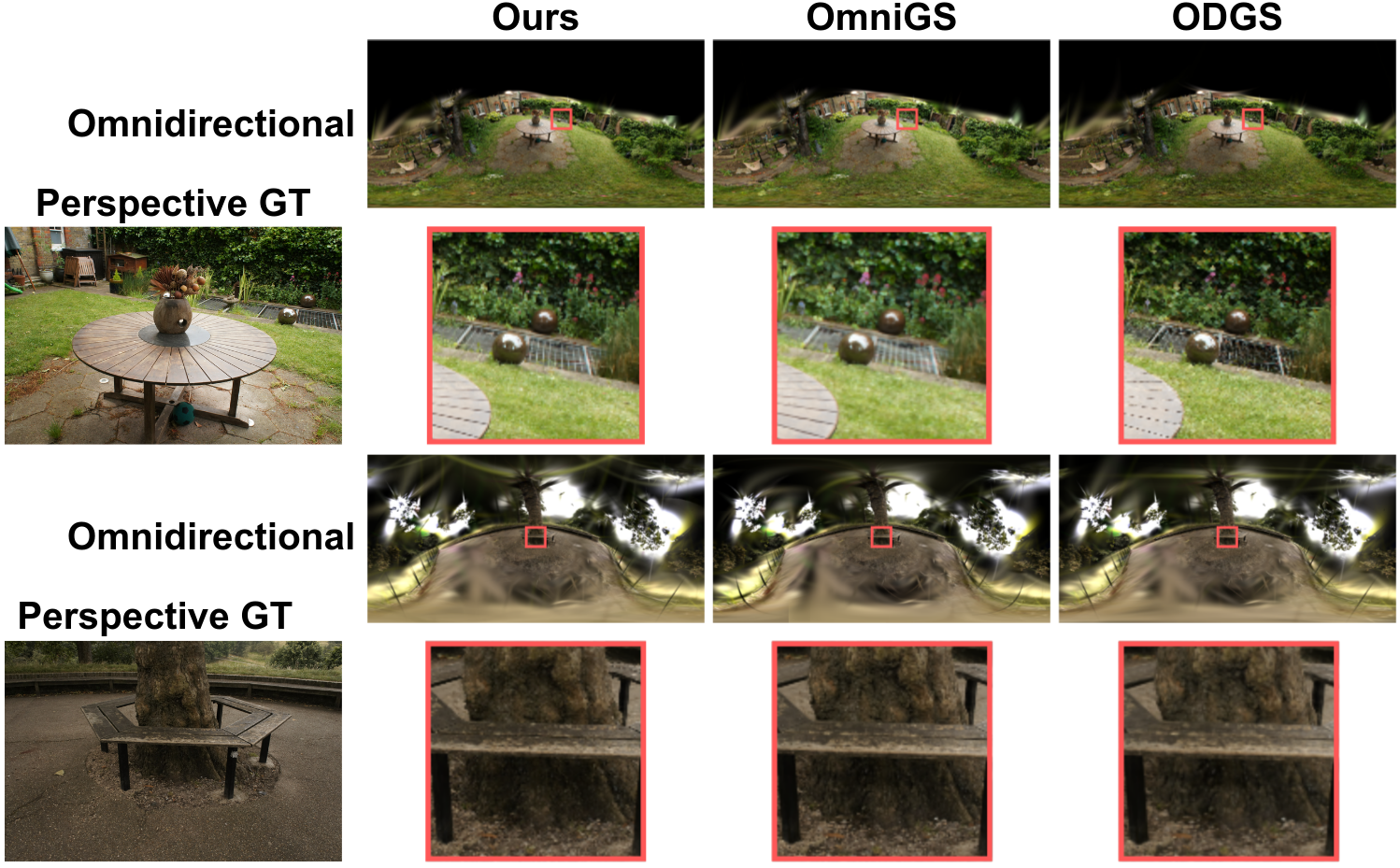}
    \caption{Omnidirectional rendering of vanilla-3DGS models trained on Mip-NeRF 360, compared with OmniGS and ODGS.}
    \label{fig:supp_exp1_2}
\end{figure}

The first rows of~\cref{fig:supp_exp1} show that UniTriSplat preserves fine details when converting perspective-trained scenes to fisheye views. The indoor examples also contain fewer floating artifacts near image boundaries. In the omnidirectional comparisons of~\cref{fig:supp_exp1_2}, UniTriSplat retains fine structures in the enlarged regions despite the incomplete spherical coverage of the original perspective training images. These results show that the HEALPix rasterizer can be applied to Gaussian scenes reconstructed by an external perspective pipeline and can synthesize camera models not used during their optimization.

\section{Implementation Details}
This section details the implementation of the UniTriSplat framework. \Cref{sec:anyfov} presents a unified camera modeling strategy that ensures compatibility across arbitrary FoV inputs, offering broader support for camera models than prior 3DGS methods. \Cref{sec:raster_detail} elaborates on the HEALPix-based rasterization, covering grid resolution scaling, dynamic memory management, and algorithmic pseudocode. Finally, \Cref{sec:hyperpara} discusses hyperparameter tuning for Gaussian optimization within the radian domain on the HEALPix grid.
\subsection{Compatibility with Any FoV Input}
\label{sec:anyfov}

Our approach accommodates perspective, fisheye, and omnidirectional projections by adopting a FoV-based camera modeling strategy, thereby enabling seamless adaptation to any input FoV. All camera models share the same Gaussian-to-HEALPix rasterization process. Only the camera-to-sphere mapping and the valid spherical region change during image synthesis. This design decouples spherical scene representation from camera-specific projection and enables perspective, fisheye, and omnidirectional rendering within a single rasterization framework.

For partial-FoV cameras, we restrict computation to HEALPix pixels inside the visible spherical region defined by the camera intrinsics and horizontal and vertical FoVs. The rendered spherical samples are then interpolated onto the target image plane. We next describe the camera-to-sphere mappings used for perspective, fisheye, and omnidirectional images.

\textbf{Perspective (Pinhole):} We adopt the standard pinhole camera model used in vanilla 3DGS. Given image dimensions $W \times H$ and field of view $\text{FoV}_x$, the focal length is $f_x = W / (2\tan(\text{FoV}_x/2))$ (similarly for $f_y$), with principal point $(c_x, c_y) = (W/2, H/2)$.

For a pixel $(u, v)$, we compute normalized coordinates $x = (u - c_x)/f_x$, $y = (v - c_y)/f_y$, then obtain the viewing direction $\mathbf{d} = (x, y, 1)^\top / \|(x, y, 1)\|$. The spherical coordinates are:
\begin{equation}
\omega = \arctan2(d_x, d_z), \quad \phi = \arcsin(d_y)
\label{eq:persp_sphere}
\end{equation}
where $\omega \in [-\pi, \pi)$ and $\phi \in [-\pi/2, \pi/2]$.

\textbf{Fisheye (Anisotropic Equidistant):} The classical equidistant fisheye model is isotropic, using a single FoV parameter $w$ such that the incident angle $\psi = w \cdot r$ where $r$ is the normalized radial distance. However, this restricts the valid region to a circle inscribed within the image, wasting pixels in rectangular sensors. We extend this to an \emph{anisotropic} equidistant model with independent horizontal and vertical FoV parameters $(w_x, w_y)$, where $w = \text{FoV}/180°$. This allows the valid region to form an ellipse that better utilizes the full image area, enabling higher effective resolution for the same sensor size.

\begin{table}[t]
\centering
\caption{Resolution hierarchy in HEALPix rendering. Given an input image of size $W\times H$ covering a solid angle $\Omega_{\mathrm{in}}$, the HEALPix resolution is selected to match the average solid angle per input pixel. The default configuration uses the nearest HEALPix order, while $k_{\mathrm{L}}$ and $k_{\mathrm{H}}$ denote the adjacent lower and higher orders used in the cross-resolution evaluation.}
\label{tab:resolution}
\setlength{\tabcolsep}{4pt}
\renewcommand{\arraystretch}{1.05}
\begin{tabular}{lcc}
\toprule
\textbf{Parameter} & \textbf{Symbol} & \textbf{Definition} \\
\midrule
Input resolution
& $W\times H$
& Image dimensions \\
Visible solid angle
& $\Omega_{\mathrm{in}}$
& Solid angle covered by the camera FoV \\
Continuous target
& $N_{\mathrm{side}}^{*}$
& $\sqrt{\dfrac{4\pi WH}{12\Omega_{\mathrm{in}}}}$ \\
Default HEALPix order
& $k$
& $\operatorname{round}\!\left(\log_2 N_{\mathrm{side}}^{*}\right)$ \\
Lower / higher orders
& $k_{\mathrm{L}},\,k_{\mathrm{H}}$
& $\left\lfloor\log_2 N_{\mathrm{side}}^{*}\right\rfloor,\,
   \left\lceil\log_2 N_{\mathrm{side}}^{*}\right\rceil$ \\
HEALPix side resolution
& $N_{\mathrm{side}}$
& $2^k$ \\
Total spherical samples
& $N_{\mathrm{pix}}$
& $12N_{\mathrm{side}}^2=12\cdot4^k$ \\
Tile order
& $k_t$
& $k-\log_2 B$ \\
Full-sphere tile count
& $N_{\mathrm{tile}}$
& $12\cdot4^{k_t}=12\left(N_{\mathrm{side}}/B\right)^2$ \\
Tile size
& $B$
& $16$ ($16\times16$ HEALPix samples) \\
\bottomrule
\end{tabular}
\end{table}

Given image dimensions $W \times H$ and requiring the valid region to be tangent to the image boundary, we set focal lengths $f_x = W/\pi$, $f_y = H/\pi$, with principal point $(c_x, c_y) = (W/2, H/2)$.

For a pixel $(u, v)$, we first compute normalized coordinates $m_x = (u - c_x)/f_x$, $m_y = (v - c_y)/f_y$, then obtain the radial distance $r_m = \sqrt{m_x^2 + m_y^2}$ and azimuth angle $\gamma = \arctan2(m_y, m_x)$, which measures the angular direction of the pixel relative to the principal point. The key to anisotropic projection is the direction-dependent FoV factor:
\begin{equation}
w_{\text{eff}}(\gamma) = \sqrt{(w_x \cos\gamma)^2 + (w_y \sin\gamma)^2}
\label{eq:fish_weff}
\end{equation}
which interpolates between $w_x$ (horizontal, $\gamma = 0$) and $w_y$ (vertical, $\gamma = \pi/2$) based on the azimuthal direction. The incident angle from the optical axis is then $\psi = r_m \cdot w_{\text{eff}}(\gamma)$, yielding the 3D viewing direction $\mathbf{d} = (\sin\psi \cos\gamma, \sin\psi \sin\gamma, \cos\psi)^\top$ in camera coordinates. Finally, the spherical coordinates are:
\begin{equation}
\omega = \arctan2(d_x, d_z), \quad \phi = \arcsin(d_y)
\label{eq:fish_sphere}
\end{equation}

The valid imaging region is defined by the following elliptical constraint:
\begin{equation}
    \frac{(u - c_x)^2}{a^2} + \frac{(v - c_y)^2}{b^2} \leq 1,
    \label{eq:imaging_region}
\end{equation}
where the semi-axes $a$ and $b$ are parameterized as:
\begin{equation}
    a = f_x \frac{\pi}{2}, \quad b = f_y \frac{\pi}{2}.
    \label{eq:semi_axes}
\end{equation}
Specifically, when $w_x = w_y = 1$,~\cref{eq:imaging_region} reduces to a standard 180$^\circ$ isotropic fisheye. For configurations where $w_x = w_y > 1$, the model accommodates super-hemispherical imaging exceeding 180$^\circ$.

\begin{algorithm}[t]
\scriptsize
\caption{HEALPix-Based 3D Gaussian Rasterization}
\label{alg:healpix_forward}
\begin{algorithmic}[1]
\Require Gaussians $\mathcal{G} = \{(\boldsymbol{\mu}_i, \mathbf{s}_i, \mathbf{q}_i, o_i, \mathbf{c}_i)\}_{i=1}^P$, view $\mathbf{V} = [\mathbf{W} \mid \mathbf{b}]$, HEALPix pixel order $k$, tile size $B = 2^b$
\Ensure HEALPix image $\mathbf{I} \in \mathbb{R}^{N_{\mathrm{pix}} \times 3}$, $N_{\mathrm{pix}} = 12 \cdot 4^k$
\State $k_t \gets k-b$ \Comment{HEALPix order of the rendering tiles}
\Statex \textbf{// Preprocessing (parallel over Gaussians)}
\For{$i = 1, \ldots, P$ \textbf{in parallel}}
    \State $\tilde{\boldsymbol{\mu}}_i \gets \mathbf{W}\boldsymbol{\mu}_i + \mathbf{b}$; \quad $(\omega_i, \phi_i) \gets \Pi(\tilde{\boldsymbol{\mu}}_i)$; \quad $\rho_i \gets \lVert\tilde{\boldsymbol{\mu}}_i\rVert_2$
    \State $\boldsymbol{\Sigma}_{3D,i} \gets \mathbf{R}_i\,\operatorname{diag}(\mathbf{s}_i^2)\,\mathbf{R}_i^\top$, where $\mathbf{R}_i = \textsc{QuatToMat}(\mathbf{q}_i)$
    \State $\mathbf{J}_i \gets \left.\dfrac{\partial \Pi(\mathbf{t})}{\partial \mathbf{t}}\right|_{\mathbf{t}=\tilde{\boldsymbol{\mu}}_i}$; \quad $\mathbf{S}_{\phi_i} \gets \operatorname{diag}(\cos\phi_i,1)$
    \State $\boldsymbol{\Sigma}_{\mathrm{arc},i} \gets \mathbf{S}_{\phi_i}\mathbf{J}_i\mathbf{W}\boldsymbol{\Sigma}_{3D,i}\mathbf{W}^\top\mathbf{J}_i^\top\mathbf{S}_{\phi_i}^\top$
    \State $r_{s,i} \gets 3\sqrt{\lambda_{\max}(\boldsymbol{\Sigma}_{\mathrm{arc},i})}$
\EndFor
\Statex \textbf{// Tile assignment and front-to-back sorting}
\For{$i = 1, \ldots, P$ \textbf{in parallel}}
    \State $\mathcal{T}_i \gets \textsc{QueryDisc}(\omega_i,\phi_i,r_{s,i},k_t)$ \Comment{NESTED or RING mode}
    \For{each $t \in \mathcal{T}_i$}
        \State Emit pair $((t,\rho_i),i)$ \Comment{Tile-depth key and Gaussian index}
    \EndFor
\EndFor
\State Sort pairs lexicographically by tile index and increasing radial depth
\State Identify per-tile ranges $[\texttt{start}_t,\texttt{end}_t)$ in the sorted Gaussian-index array
\Statex \textbf{// Parallel rendering}
\For{each pixel $p \in \{0,\ldots,N_{\mathrm{pix}}-1\}$ \textbf{in parallel}}
    \State $(\omega_p,\phi_p) \gets \textsc{Nest2Lonlat}(p,k)$; \quad $t \gets \textsc{AncestorTile}(p,k_t)$
    \State $T \gets 1$; \quad $\mathbf{C} \gets \mathbf{0}$
    \For{$j = \texttt{start}_t$ \textbf{to} $\texttt{end}_t-1$}
        \State $i \gets \texttt{gaussian\_id}[j]$
        \State $\mathbf{d} \gets \textsc{SphericalOffset}((\omega_p,\phi_p),(\omega_i,\phi_i))$
        \State $\alpha_i \gets o_i\exp\!\left(-\tfrac{1}{2}\mathbf{d}^\top\boldsymbol{\Sigma}_{\mathrm{arc},i}^{-1}\mathbf{d}\right)$
        \State $\mathbf{C} \gets \mathbf{C}+T\alpha_i\mathbf{c}_i$; \quad $T \gets T(1-\alpha_i)$
        \If{$T < \epsilon$} \State \textbf{break} \EndIf
    \EndFor
    \State $\mathbf{I}[p] \gets \mathbf{C}$
\EndFor
\end{algorithmic}
\end{algorithm}

\textbf{Omnidirectional:} For a full $360^\circ \times 180^\circ$ panoramic image of size $W \times H$, each pixel $(u, v)$ maps directly to spherical coordinates $\omega = 2\pi u/W - \pi$ and $\phi = \pi/2 - \pi v/H$, yielding direction $\mathbf{d} = (\cos\phi \sin\omega, \sin\phi, \cos\phi \cos\omega)^\top$.

\textbf{Unified Spherical Rendering:} Regardless of the input camera model, all pixels are first projected to spherical coordinates $(\omega, \phi)$ using the respective projection equations described above. Our method then renders directly in HEALPix space, which provides an equal-area tessellation of the sphere with $N_{\text{pix}} = 12 \times N_{\text{side}}^2$ pixels covering the full $4\pi$ steradians. This unified spherical representation enables seamless handling of arbitrary FoV inputs within a single rendering framework.

\begin{algorithm}[t]
\scriptsize
\caption{\textsc{QueryDiscNested}: Quadtree Traversal}
\label{alg:nested_query}
\begin{algorithmic}[1]
\Require Disc center $(\omega, \phi)$, radius $r_s$, tile order $k$, refinement $f$
\Ensure Overlapping tile indices $\mathcal{T}$
\State $k_{\max} \gets k + \log_2 f$; \quad $\texttt{stack} \gets \{(p, 0) : p \in [0,11]\}$; \quad $\mathcal{T} \gets \emptyset$
\While{$\texttt{stack} \neq \emptyset$}
    \State Pop $(p, o)$; \quad $d \gets \textsc{Haversine}((\omega,\phi), \textsc{Pix2Lonlat}(p,o))$
    \State $z \gets \textsc{Classify}(d, r_s, \textsc{PixRad}(o))$ \Comment{0:out, 1:partial, 2+:overlap}
    \If{$z = 0$} \textbf{continue} \Comment{Prune}
    \ElsIf{$z \geq 2$ \textbf{and} $o \geq k$} $\mathcal{T} \gets \mathcal{T} \cup \{\lfloor p/4^{o-k} \rfloor\}$
    \ElsIf{$o < k_{\max}$} Push $(4p+c, o+1)$ for $c \in \{0,1,2,3\}$
    \Else\ $\mathcal{T} \gets \mathcal{T} \cup \{\lfloor p/4^{o-k} \rfloor\}$
    \EndIf
\EndWhile
\State \Return $\mathcal{T}$
\end{algorithmic}
\end{algorithm}

\begin{algorithm}[h]
\scriptsize
\caption{\textsc{QueryDiscRing}: RING Sequential Scan}
\label{alg:ring_query}
\begin{algorithmic}[1]
\Require Disc center $(\omega,\phi)$, radius $r_s$, tile order $k$
\Ensure Overlapping tile indices $\mathcal{T}$
\State $N_{\mathrm{side}} \gets 2^k$; \quad $z_c \gets \sin\phi$; \quad $\mathcal{T} \gets \emptyset$
\State $\phi_{\min} \gets \max(-\pi/2,\phi-r_s)$; \quad $\phi_{\max} \gets \min(\pi/2,\phi+r_s)$
\State $z_{\min} \gets \sin\phi_{\min}$; \quad $z_{\max} \gets \sin\phi_{\max}$
\State $(i_{r,\min},i_{r,\max}) \gets \textsc{ZtoRingRange}(z_{\min},z_{\max},N_{\mathrm{side}})$
\For{$i_r = i_{r,\min}$ \textbf{to} $i_{r,\max}$}
    \State $(z_r,n_r,\omega_0) \gets \textsc{RingInfo}(i_r)$
    \State $\delta \gets \sqrt{\max(0,1-z_c^2)}\sqrt{\max(0,1-z_r^2)}$
    \If{$\delta < \epsilon$} \Comment{Disc center or ring at a pole}
        \If{$z_cz_r \geq \cos r_s$}
            \State Add all $n_r$ pixels in ring $i_r$ to $\mathcal{T}$
        \EndIf
        \State \textbf{continue}
    \EndIf
    \State $c_\omega \gets (\cos r_s-z_cz_r)/\delta$
    \If{$c_\omega \leq -1$}
        \State Add all $n_r$ pixels in ring $i_r$ to $\mathcal{T}$ \Comment{Full ring}
    \ElsIf{$c_\omega < 1$}
        \State $\Delta\omega \gets \arccos\!\left(\operatorname{clamp}(c_\omega,-1,1)\right)$
        \State $i_{p,\min} \gets \left\lfloor\dfrac{(\omega-\Delta\omega-\omega_0)n_r}{2\pi}\right\rfloor$
        \State $i_{p,\max} \gets \left\lceil\dfrac{(\omega+\Delta\omega-\omega_0)n_r}{2\pi}\right\rceil$
        \For{$i_p = i_{p,\min}$ \textbf{to} $i_{p,\max}$}
            \State $\mathcal{T} \gets \mathcal{T}\cup\{\textsc{Ring2Nest}(i_r,\operatorname{mod}(i_p,n_r))\}$
        \EndFor
    \EndIf
\EndFor
\State \Return $\mathcal{T}$
\end{algorithmic}
\end{algorithm}

\subsection{Implementation of the HEALPix-based Rasterization}
\label{sec:raster_detail}

\Cref{alg:healpix_forward} shows the pipeline of the HEALPix-based 3D Gaussian rasterization. The forward pass follows the three-stage pipeline of standard 3DGS, adapted for HEALPix grids. Stage 1 (preprocessing) transforms Gaussians to camera space and computes arc-length covariances via~\cref{eq:arc_scaling} as described in~\cref{sec:spherical_projection}. Stage 3 (rendering) performs per-pixel alpha compositing using great-circle distances per~\cref{eq:unified_rendering}. We focus here on Stage 2: tile assignment strategies.

\paragraph{Stage 2: Tile Query Methods.}
We provide two tile query algorithms exploiting HEALPix's dual indexing schemes.

\textbf{NESTED Quadtree Traversal.}
\Cref{alg:nested_query} shows the pipeline of the NESTED quadtree traversal. We adapt the classical query-pixel algorithm from the HEALPix~\cite{healpix}, which performs depth-first search (DFS) through HEALPix's hierarchical NESTED structure. Starting from the 12 base quadrilaterals, each node is classified into zones based on its distance to the query disc center: zone 3 (fully inside), zone 2 (center inside), zone 1 (partially overlapping), or zone 0 (fully outside). Subtrees in zone 0 are pruned; zones 2--3 are added to results; zone 1 nodes are recursively subdivided.

To integrate this into our GPU rasterization pipeline, we search at a finer resolution $N_{\text{side}}^{\text{fact}} = \text{fact} \cdot N_{\text{side}}$ (where $\text{fact}$ is typically 4 or 8) to handle boundary cases precisely, then map results back to tile resolution. This achieves $O(K + \log N_{\text{side}}^{\text{fact}})$ time per Gaussian but requires $O(\log N_{\text{side}}^{\text{fact}})$ stack memory per thread.

\textbf{RING Sequential Scan.}
\Cref{alg:ring_query} shows the pipeline of the RING sequential scan. This method exploits HEALPix's RING ordering, which arranges pixels along iso-latitude rings characterized by $z = \cos\vartheta$ where $\vartheta$ is the colatitude. Given a spherical disc with center $(\omega_c, \phi_c)$ and radius $r_s$:
\begin{enumerate}[leftmargin=*, nosep]
    \item Compute $z$-bounds: $z_{\min} = \cos(\phi_c - r_s + \frac{\pi}{2})$, $z_{\max} = \cos(\phi_c + r_s + \frac{\pi}{2})$
    \item Convert to ring indices $i_{r,\min}$, $i_{r,\max}$ using HEALPix ring formulas
    \item For each ring $i_r$, analytically solve for the longitude half-width via~\cref{eq:ring_phi_range}
    \item Enumerate pixels within the $(\omega_c - \Delta\omega, \omega_c + \Delta\omega)$ range on each ring
\end{enumerate}
This achieves $O(K)$ time with $O(1)$ memory and better GPU cache coherence due to sequential ring access. RING offers approximately $1.7\times$ speedup over NESTED with marginal quality loss, as detailed in~\cref{sec:ablation}.

\subsection{Training Parameter Scaling}
\label{sec:hyperpara}

The gradient dynamics of our method differ from the pinhole model due to the discrepancy in the projection domain. Specifically, the pinhole projection Jacobian scales with the focal length $f$ as $\mathbf{J}_{\text{pixel}} \propto f/z$, whereas the spherical counterpart is governed by $\mathbf{J}_{\text{rad}} \propto 1/r$. Given that typical focal lengths $f$ range from $500$ to $1000$, the spherical Jacobian magnitude is attenuated by several orders of magnitude. According to the chain rule, $\partial \mathcal{L}/\partial \boldsymbol{\mu}_{3D} = (\partial \mathcal{L}/\partial \boldsymbol{\mu}_{2D}) \cdot \mathbf{J}$, this reduction directly suppresses the 3D position gradients. To compensate for this gradient decay and ensure robust convergence, we scale the position learning rate upward by approximately one order of magnitude.

Conversely, the densification threshold—which governs adaptive Gaussian splitting—requires a downward recalibration to account for reduced gradient accumulation in the HEALPix domain. This phenomenon stems from three primary factors: (i) sparse effective sampling when the HEALPix resolution is coarser than the input imagery; (ii) diminished per-Gaussian activation frequency in scenarios where training views provide only partial spherical coverage; and (iii) the $\cos^2\phi$ metric distortion that attenuates gradient contributions near the poles. Consequently, we lower the densification threshold to maintain a splitting sensitivity consistent with vanilla 3DGS.

\bibliographystylesupp{splncs04}
\bibliographysupp{supp}

\vspace{2em}

\end{document}